\PassOptionsToPackage{dvipsnames,table,xcdraw}{xcolor}
\documentclass{paper}
\usepackage{amssymb}
\usepackage{paper-support}
\usepackage{paper-figures}
\usepackage{preprint-layout}
\ifdefined\PFFExperimentConstantsLoaded
  \let\PFFExperimentConstantsLoadAction 
\else
  \let\PFFExperimentConstantsLoadAction\relax
\fi
\PFFExperimentConstantsLoadAction
\def\PFFExperimentConstantsLoaded{1}

\def\PFFExperimentNoControlReportingMNmRoundedOne{88.3}

\ifdefined\PFFProtocolConstantsLoaded
  \let\PFFProtocolConstantsLoadAction 
\else
  \let\PFFProtocolConstantsLoadAction\relax
\fi
\PFFProtocolConstantsLoadAction
\def\PFFProtocolConstantsLoaded{1}

\def\PFFProtocolSampleRateHz{30}
\def\PFFProtocolEpisodeDurationS{60}

\def\PFFProtocolSamplesPerEpisodeDisplay{1{,}800}
\def\PFFProtocolTrainingBudgetMinutes{60}
\def\PFFProtocolTrainingEpisodes{60}

\def\PFFProtocolTrainingStepsDisplay{108{,}000}
\def\PFFProtocolMainRepetitions{10}
\def\PFFProtocolEndpointEpisodeCount{10}
\def\PFFProtocolEndpointFirstIndex{50}
\def\PFFProtocolEndpointLastIndex{59}
\def\PFFProtocolStabilizationS{60}
\def\PFFProtocolIncomingFlowSpeedMPerS{0.12}
\def\PFFProtocolAquariumLengthM{0.90}
\def\PFFProtocolAquariumWidthM{0.30}
\def\PFFProtocolAquariumHeightM{0.30}
\def\PFFProtocolAquariumNominalVolumeL{90}
\def\PFFProtocolTestSectionLengthM{0.15}
\def\PFFProtocolTestSectionWidthM{0.10}
\def\PFFProtocolTestSectionHeightM{0.15}
\def\PFFProtocolPropellerCount{3}
\def\PFFProtocolCylinderRadiusM{0.021}

\def\PFFProtocolEncoderPPR{640}
\def\PFFProtocolMaximumRotationRateRadPerS{15.708}
\def\PFFProtocolMaximumSurfaceSpeedMPerSRoundedTwo{0.33}
\def\PFFProtocolStrouhalUpperBound{0.50}
\def\PFFProtocolInflowSpeedUpperBoundMPerS{0.2}
\def\PFFProtocolControlCharacteristicFrequencyHzRoundedOne{2.4}
\def\PFFProtocolReynoldsNumberDisplay{5.0\times10^{3}}
\def\PFFProtocolBlockageRatioRoundedTwo{0.42}
\def\PFFProtocolTorqueSensorFullScaleMNm{150}
\def\PFFProtocolTorqueSensorAccuracyPercentFullScale{0.1}

\def\PFFProtocolTorqueSampleRateHzDisplay{1{,}000}
\def\PFFProtocolTorqueFilterWindowS{1}

\def\PFFProtocolTorqueSensorUncertaintyUpperMNmRoundedOne{0.2}
\def\PFFProtocolFabricationAssemblyUncertaintyUpperMNm{0.6}
\def\PFFProtocolMaterialEffectThresholdMNm{1.0}
\def\PFFProtocolPIVCameraRateHz{60}
\def\PFFProtocolPIVTracerDiameterUm{10}
\def\PFFProtocolPIVCroppedImageMegapixels{1}
\def\PFFProtocolPIVProcessingGridRows{512}
\def\PFFProtocolPIVProcessingGridColumns{512}
\def\PFFProtocolFlowObservationGridRows{16}
\def\PFFProtocolFlowObservationGridColumns{16}

\def\PFFProtocolActionNormalizedLowerBound{-1}
\def\PFFProtocolActionNormalizedUpperBound{1}
\def\PFFProtocolComputeGPUCount{4}
\def\PFFProtocolConfidenceLevelPercent{95}

\def\PFFProtocolBootstrapResamplesDisplay{100{,}000}

\def\PFFProtocolReferenceAttainmentPercent{80}

\def\PFFProtocolStudentTQuantileProbability{0.975}

\def\PFFProtocolReplaySeeds{3}
\def\PFFProtocolReplayCheckpoints{60}
\def\PFFProtocolReplayRepeats{5}

\def\PFFProtocolReplayTrajectoriesDisplay{1{,}800}
\def\PFFProtocolReplayStudentTDegreesFreedom{2}
\def\PFFProtocolDreamerAdditionalConfigurationsPerObjective{6}
\def\PFFProtocolDreamerAdditionalRepetitions{3}
\def\PFFProtocolDreamerModelSizeRepetitions{3}
\def\PFFProtocolDreamerZeroedFlowRepetitions{3}
\def\PFFProtocolBaselineNoControlAmplitudeRadPerS{0.000}
\def\PFFProtocolBaselineGridAmplitudesRadPerS{7.854,10.472,13.090,15.708}
\def\PFFProtocolBaselineConstantAmplitudesRadPerS{-15.708,-13.090,-10.472,-7.854,0.000,7.854,10.472,13.090,15.708}
\def\PFFProtocolBaselineGridFrequencyMinHz{0.0}
\def\PFFProtocolBaselinePositiveFrequencyMinHz{0.1}
\def\PFFProtocolBaselineGridFrequencyMaxHz{3.0}
\def\PFFProtocolBaselineGridFrequencyStepHz{0.1}

\def\PFFProtocolBaselinePositiveFrequencyConditions{30}
\def\PFFProtocolBaselineGridSettings{129}
\def\PFFProtocolBaselineGridRepetitions{3}
\def\PFFProtocolBaselineGridPlateauOnsetHz{2.5}
\def\PFFProtocolBaselineGridSelectedMaxFrequencyHz{0.8}
\def\PFFProtocolBaselineGridSelectedMinFrequencyHz{3.0}
\def\PFFProtocolBaselineMotorCharacterizationRepetitions{3}
\def\PFFProtocolBaselineStudentTDegreesFreedom{2}
\def\PFFProtocolBaselineSelectedAmplitudeRadPerS{15.708}
\def\PFFProtocolBaselineMinimumControlUpdatesPerPeriod{10}

\def\PFFProtocolTemporalPrimarySegmentSeconds{20}

\def\PFFProtocolTemporalWelchSegments{5}

\def\PFFProtocolTemporalMotorFeedbackSpikeThresholdRPM{500}

\def\PFFProtocolInverseInitialSearchWindowS{5}
\def\PFFProtocolInverseLocalAverageWindowS{0.5}

\def\PFFProtocolInverseSustainedResponseWindowS{10}

\def\PFFProtocolInverseSustainedDragIncreaseReferencePercent{50}
\def\PFFProtocolAlgorithmRepetitions{3}
\def\PFFProtocolAlgorithmEpisodes{30}
\def\PFFProtocolAlgorithmBudgetMinutes{30}

\def\PFFProtocolAlgorithmLibraryVersion{2.7.0}
\def\PFFProtocolAlgorithmDiscountFactor{0.999}
\def\PFFProtocolExtendedSACHours{10}

\def\PFFProtocolStandardModelParametersMillions{25}
\def\PFFProtocolSmallModelParametersMillions{1}

\ifdefined\PFFPaperNumbersLoaded
  \expandafter 
\fi
\def\PFFPaperNumbersLoaded{1}

\def\PFFPaperHeadlineMaxAttainmentDragChangePercent{21.3}

\def\PFFPaperHeadlineMinAttainmentDragChangePercent{23.7}

\def\PFFPaperHeadlineMaxFlowEndpointPercentMean{25.5}
\def\PFFPaperHeadlineMaxFlowEndpointPercentSD{0.9}
\def\PFFPaperHeadlineMaxFlowEndpointPercentPM{25.5\mathbin{\pm}0.9}

\def\PFFPaperHeadlineMaxNoFlowEndpointPercentMean{1.8}
\def\PFFPaperHeadlineMaxNoFlowEndpointPercentSD{4.3}
\def\PFFPaperHeadlineMaxNoFlowEndpointPercentPM{1.8\mathbin{\pm}4.3}

\def\PFFPaperHeadlineMinFlowEndpointPercentMean{32.4}
\def\PFFPaperHeadlineMinFlowEndpointPercentSD{1.6}
\def\PFFPaperHeadlineMinFlowEndpointPercentPM{32.4\mathbin{\pm}1.6}

\def\PFFPaperHeadlineMinNoFlowEndpointPercentMean{31.4}
\def\PFFPaperHeadlineMinNoFlowEndpointPercentSD{2.1}
\def\PFFPaperHeadlineMinNoFlowEndpointPercentPM{31.4\mathbin{\pm}2.1}

\def\PFFPaperReplayMaxConfidenceLowerMNm{0.2}
\def\PFFPaperReplayMaxConfidenceUpperMNm{5.0}

\def\PFFPaperReplayMinConfidenceLowerMNm{-5.3}
\def\PFFPaperReplayMinConfidenceUpperMNm{5.1}

\def\PFFPaperObservationMaxDifferenceMNm{21.4}
\def\PFFPaperObservationMaxDifferencePercentagePoints{24.2}
\def\PFFPaperObservationMaxConfidenceLowerMNm{16.4}
\def\PFFPaperObservationMaxConfidenceUpperMNm{26.9}
\def\PFFPaperObservationMinDifferenceMNm{3.1}
\def\PFFPaperObservationMinDifferencePercentagePoints{3.5}
\def\PFFPaperObservationMinConfidenceLowerMNm{0.5}
\def\PFFPaperObservationMinConfidenceUpperMNm{6.9}
\def\PFFPaperModelSizeMaxDifferenceMNm{-9.4}
\def\PFFPaperModelSizeMaxDifferencePercentagePoints{-10.7}
\def\PFFPaperModelSizeMaxConfidenceLowerMNm{-11.5}
\def\PFFPaperModelSizeMaxConfidenceUpperMNm{-6.9}
\def\PFFPaperFlowRemovalMaxDifferenceMNm{1.9}
\def\PFFPaperFlowRemovalMaxDifferencePercentagePoints{2.1}
\def\PFFPaperFlowRemovalMaxConfidenceLowerMNm{-6.7}
\def\PFFPaperFlowRemovalMaxConfidenceUpperMNm{10.2}
\def\PFFPaperObservationMaxFlowRunCount{19}

\def\PFFPaperObservationMaxNoFlowRunCount{19}
\def\PFFPaperObservationMaxNoFlowAttainedCount{1}

\def\PFFPaperObservationMaxRestrictedTimeDifferenceMinutes{49.7}
\def\PFFPaperObservationMaxRestrictedTimeConfidenceLowerMinutes{43.2}
\def\PFFPaperObservationMaxRestrictedTimeConfidenceUpperMinutes{53.8}
\def\PFFPaperObservationMaxDispersionDifferenceMNm{2.23}
\def\PFFPaperObservationMaxDispersionConfidenceLowerMNm{-0.04}
\def\PFFPaperObservationMaxDispersionConfidenceUpperMNm{4.59}
\def\PFFPaperObservationMinFlowRunCount{19}

\def\PFFPaperObservationMinNoFlowRunCount{19}

\def\PFFPaperObservationMinNoFlowFinalWindowAttainedCount{16}

\def\PFFPaperObservationMinRestrictedTimeDifferenceMinutes{16.7}
\def\PFFPaperObservationMinRestrictedTimeConfidenceLowerMinutes{3.1}
\def\PFFPaperObservationMinRestrictedTimeConfidenceUpperMinutes{32.5}
\def\PFFPaperObservationMinDispersionDifferenceMNm{2.89}
\def\PFFPaperObservationMinDispersionConfidenceLowerMNm{0.19}
\def\PFFPaperObservationMinDispersionConfidenceUpperMNm{4.17}
\def\PFFPaperZeroedFlowRunCount{3}
\def\PFFPaperZeroedFlowAttainedCount{1}
\def\PFFPaperZeroedFlowFinalWindowAttainedCount{0}
\def\PFFPaperZeroedFlowFirstAttainmentMinute{57}

\def\PFFPaperReplayMaxDifferenceMNmPM{2.6\mathbin{\pm}1.0}

\def\PFFPaperReplayMaxAllCheckpointPercentPM{19.3\mathbin{\pm}1.6}

\def\PFFPaperReplayMaxEndpointPercentMean{23.2}
\def\PFFPaperReplayMaxEndpointPercentSD{2.2}
\def\PFFPaperReplayMaxEndpointPercentPM{23.2\mathbin{\pm}2.2}
\def\PFFPaperReplayMinDifferenceMNmPM{-0.1\mathbin{\pm}2.1}

\def\PFFPaperReplayMinAllCheckpointPercentPM{30.8\mathbin{\pm}1.8}

\def\PFFPaperReplayMinEndpointPercentMean{32.1}
\def\PFFPaperReplayMinEndpointPercentSD{3.0}
\def\PFFPaperReplayMinEndpointPercentPM{32.1\mathbin{\pm}3.0}

\def\PFFPaperTemporalOnlineTrajectories{360}

\def\PFFPaperTemporalCommandNumericalConversionShiftRPM{1}
\def\PFFPaperTemporalCommandSpectrumDivergenceUpperBoundBits{2\times10^{-7}}
\def\PFFPaperTemporalMotorFeedbackOnlineInterpolatedSamples{12}
\def\PFFPaperTemporalMotorFeedbackOnlineAffectedTrajectories{12}
\def\PFFPaperTemporalMotorFeedbackReplayInterpolatedSamples{89}
\def\PFFPaperTemporalMotorFeedbackReplayAffectedTrajectories{82}
\def\PFFPaperTemporalMaxMotorUnmodifiedSpectrumDivergenceBitsMean{0.01682}
\def\PFFPaperTemporalMaxMotorUnmodifiedReplayDispersionBitsMean{0.01583}
\def\PFFPaperTemporalMaxMotorInterpolatedSpectrumDivergenceBitsMean{0.000855}
\def\PFFPaperTemporalMaxMotorInterpolatedReplayDispersionBitsMean{0.000867}

\def\PFFPaperTemporalMinMotorUnmodifiedSpectrumDivergenceBitsMean{0.01955}
\def\PFFPaperTemporalMinMotorUnmodifiedReplayDispersionBitsMean{0.02192}
\def\PFFPaperTemporalMinMotorInterpolatedSpectrumDivergenceBitsMean{0.000704}
\def\PFFPaperTemporalMinMotorInterpolatedReplayDispersionBitsMean{0.000641}

\def\PFFPaperBaselineStudentTCriticalValue{4.303}

\def\PFFPaperBaselineMaxReferencePercentMean{26.6}
\def\PFFPaperBaselineMaxReferencePercentSD{0.7}
\def\PFFPaperBaselineMaxReferencePercentPM{26.6\mathbin{\pm}0.7}
\def\PFFPaperBaselineMaxReferenceConfidenceLowerPercent{24.9}
\def\PFFPaperBaselineMaxReferenceConfidenceUpperPercent{28.3}

\def\PFFPaperBaselineMinReferencePercentMean{29.7}
\def\PFFPaperBaselineMinReferencePercentSD{1.3}
\def\PFFPaperBaselineMinReferencePercentPM{29.7\mathbin{\pm}1.3}
\def\PFFPaperBaselineMinReferenceConfidenceLowerPercent{26.4}
\def\PFFPaperBaselineMinReferenceConfidenceUpperPercent{32.9}

\def\PFFPaperBaselineGridNumericalMinimumFrequencyHz{2.9}

\def\PFFPaperMotorImpactMinimumMNm{-0.5}
\def\PFFPaperMotorImpactMaximumMNm{0.6}

\def\PFFPaperMotorImpactSpanPercent{1.2}

\def\PFFPaperAccountingDreamerPartialObservationAblationSamples{276}

\def\PFFPaperNegativeOnlineEligible{4918}

\def\PFFPaperNegativeOnlineMaterialPercent{94.4}
\def\PFFPaperNegativeOnlineSustainedPositive{1092}

\def\PFFPaperNegativeOnlineSustainedPositivePrecededByMaterialNegativePercent{96.1}
\def\PFFPaperNegativePairedReplayEligible{360}

\def\PFFPaperNegativePairedReplaySustainedPositive{161}

\def\PFFPaperNegativeSinusoidalWithFlowEligible{129}

\def\PFFPaperNegativeSinusoidalWithFlowMaterialPercent{71.3}

\def\PFFPaperNegativeSinusoidalHigherFrequencyEligible{11}

\def\PFFPaperNegativeSinusoidalWithoutFlowEligible{129}

\def\PFFPaperNegativeZeroedFlowEligible{180}

\def\PFFPaperNegativeZeroedFlowMaterialPercent{88.3}

\def\PFFPaperNegativeSmallModelEligible{180}

\def\PFFPaperNegativeAlgorithmBaselinesEligible{270}

\def\PFFPaperNegativeAlgorithmBaselinesMaterialPercent{94.8}

\def\PFFPaperMaterialEffectThresholdPercent{1.1}

\def\PFFPaperInverseSustainedDragIncreaseThresholdPercent{13.3}



\title{Privileged observations enable rapid and reliable policy discovery directly in the physical world}
\runningtitle{Privileged observations for physical policy discovery}
\author{Antonio Terpin\quad and\quad Raffaello D'Andrea\\[6pt]
  {\small ETH Z\"urich, Z\"urich, Switzerland}\\[2pt]
  {\small\href{mailto:aterpin@ethz.ch}{\texttt{aterpin@ethz.ch}}\quad
  \href{mailto:rdandrea@ethz.ch}{\texttt{rdandrea@ethz.ch}}}}
\date{Preprint}
\logofile{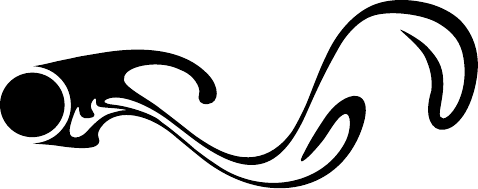}
\newcommand{\paperavailabilityfootnote}[1]{}

\begin{document}
\maketitle
\begin{abstract}
  
\noindent
We study how privileged information about a physical system affects the discovery of high-performing policies when training a reinforcement learning agent directly in the physical world.
We let the agent control a cylinder in a tabletop water channel to maximize or minimize drag. The flow is chaotic and difficult to model or simulate accurately and good strategies are not obvious beforehand.
Decades-old experimental studies provide recipes for simple, high-performance, periodic open-loop policies that increase drag by $\textnormal{\PFFPaperBaselineMaxReferencePercentMean}\mathnormal{\%}\mathbin{\pm}\textnormal{\PFFPaperBaselineMaxReferencePercentSD}\mathnormal{\%}$ and reduce it by $\textnormal{\PFFPaperBaselineMinReferencePercentMean}\mathnormal{\%}\mathbin{\pm}\textnormal{\PFFPaperBaselineMinReferencePercentSD}\mathnormal{\%}$. With dense observations of the cylinder wake, the agent learns within tens of minutes policies that increase drag by $\textnormal{\PFFPaperHeadlineMaxFlowEndpointPercentMean}\mathnormal{\%}\mathbin{\pm}\textnormal{\PFFPaperHeadlineMaxFlowEndpointPercentSD}\mathnormal{\%}$ and reduce it by $\textnormal{\PFFPaperHeadlineMinFlowEndpointPercentMean}\mathnormal{\%}\mathbin{\pm}\textnormal{\PFFPaperHeadlineMinFlowEndpointPercentSD}\mathnormal{\%}$. We record action trajectories during online policy execution and replay them open loop as fixed sequences; these replays increase drag by $\textnormal{\PFFPaperReplayMaxEndpointPercentMean}\mathnormal{\%}\mathbin{\pm}\textnormal{\PFFPaperReplayMaxEndpointPercentSD}\mathnormal{\%}$ and reduce it by $\textnormal{\PFFPaperReplayMinEndpointPercentMean}\mathnormal{\%}\mathbin{\pm}\textnormal{\PFFPaperReplayMinEndpointPercentSD}\mathnormal{\%}$. However, when we withhold flow observations during training, the agent still learns to decrease drag by $\textnormal{\PFFPaperHeadlineMinNoFlowEndpointPercentMean}\mathnormal{\%}\mathbin{\pm}\textnormal{\PFFPaperHeadlineMinNoFlowEndpointPercentSD}\mathnormal{\%}$, but no run learns to increase it ($\textnormal{\PFFPaperHeadlineMaxNoFlowEndpointPercentMean}\mathnormal{\%}\mathbin{\pm}\textnormal{\PFFPaperHeadlineMaxNoFlowEndpointPercentSD}\mathnormal{\%}$ drag increase).
Our physical experiments demonstrate an extreme case: privileged observations can be decisive for policy discovery even when the resulting actions can be replayed in open loop.
\providecommand{\paperavailabilityfootnote}[1]{\footnote{#1}}
\paperavailabilityfootnote{The designs, source code, and experimental data are available at \url{https://activefluidcontrol.com}.}

\end{abstract}
\PFFFigure[.90\linewidth]{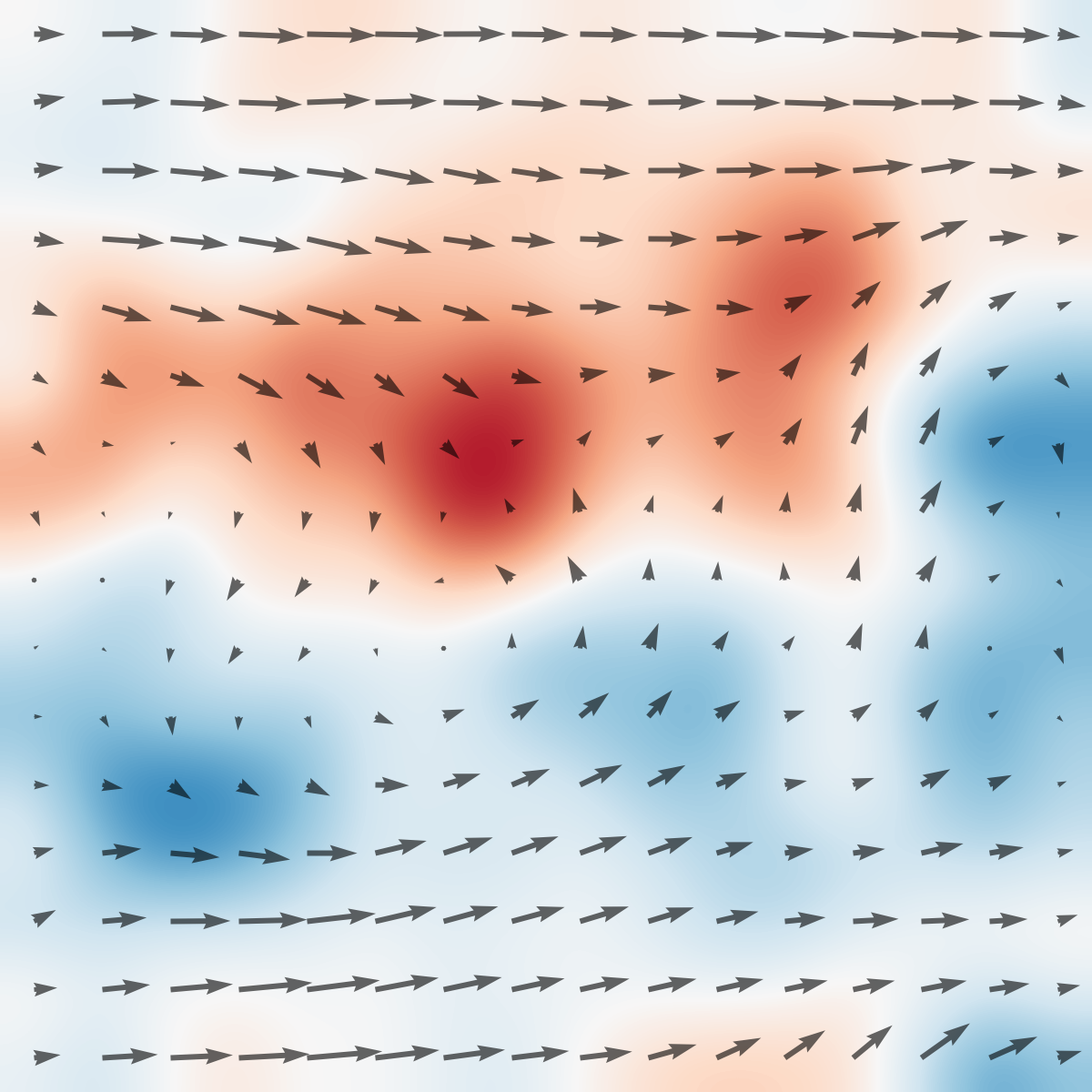}
\begin{center}
  \small Designs, source code, and experimental data:\quad
  \href{https://activefluidcontrol.com}{\textsf{activefluidcontrol.com}}
\end{center}
\clearpage
\glsresetall
\section{Introduction}
\label{sec:introduction}

People acquire many skills using information that is unavailable or presented differently when they put those skills into practice. A figure skater can draw on a coach and video during practice, a pitcher on a radar gun and repeated trials, and a barista on close inspection of each pour. Developmental studies similarly trace how visual information contributes to postural control and locomotion as balance and movement change with experience \citep{lee1974visual,anderson2019balance}. These examples make a simple distinction vivid: people can benefit from an observation while refining a skill without relying on it in the same way during later performance.

To see how an observation can play a different role within a single task, consider a simple game.
First, you crumple a scrap of paper into a ball and throw it a few meters away. You look carefully to note where it lands, then close your eyes and walk toward where you think it is, trying to pick it up without peeking.
See \cref{fig:intuitive-game}.

\PFFFigure[.65\linewidth]{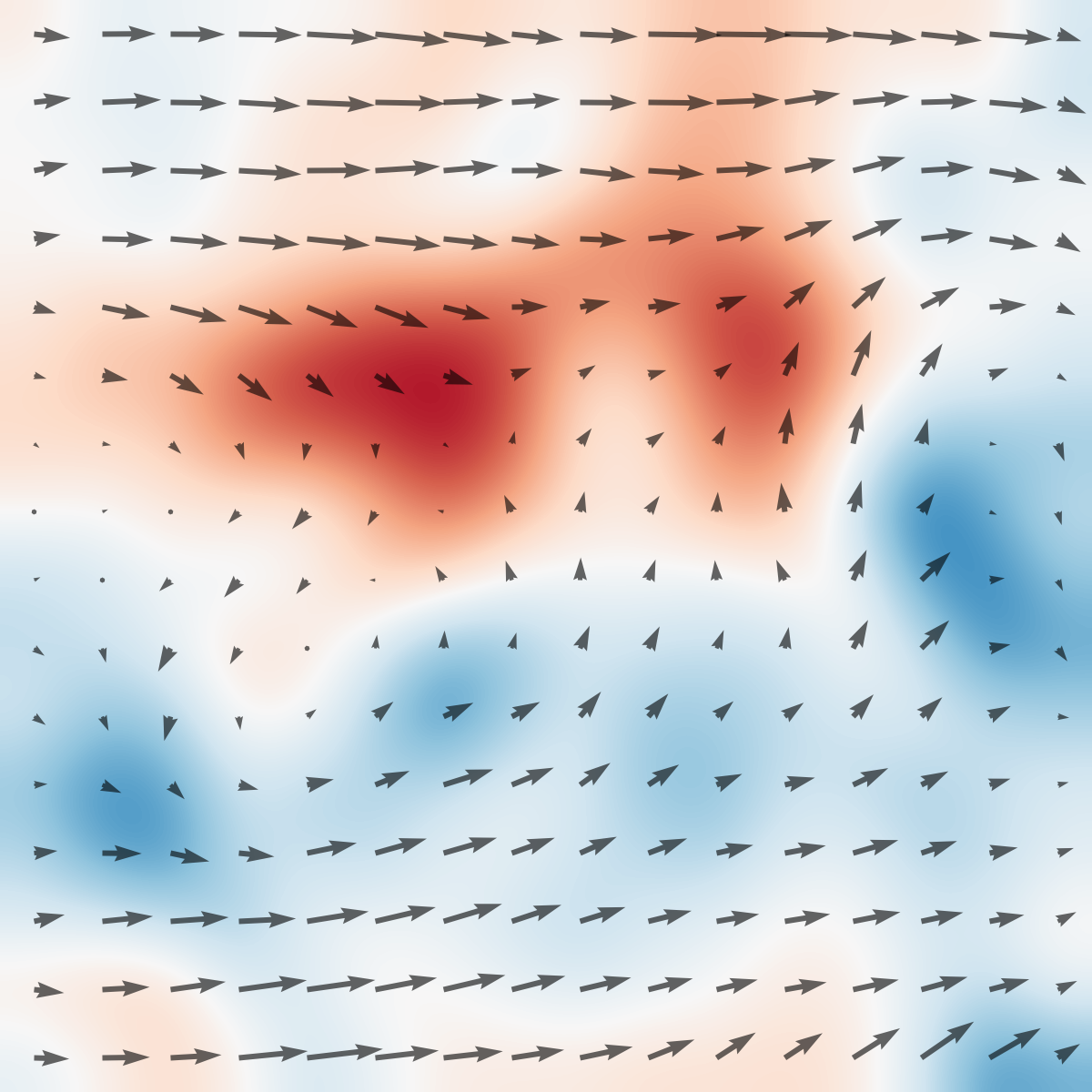}

When you reach down, you will often find that you are just a little off. Now suppose that you may peek once: that glimpse is usually enough to resolve the remaining error. You reached the neighborhood of the target using the information still available from your body and surroundings, and vision then served a more precise corrective role. Blind-walking studies show the striking accuracy of such behavior \citep{iosa2012walking} and its systematic dependence on the available visual and environmental information \citep{souman2009walking}. A related treadmill study reports lighting-dependent changes in gait stability \citep{naaman2023young}. The game illustrates the narrow intuition we need: different observations can contribute differently at different stages of a task.

Motor-control research offers several complementary ways to organize this intuition. Closed-loop accounts emphasize error-correcting feedback \citep{adams1971closed}, while schema- and program-based accounts describe how experience supports parameterized classes of movement \citep{schmidt1975schema}. A critical review found only limited support for schema theory's variability-of-practice prediction \citep{vanrossum1990schema}. Perception--action and ecological accounts emphasize continuous coupling to task-relevant information and the education of attention \citep{lee1974visual,jacobs2007direct}. Bernstein's degrees-of-freedom and motor-equivalence problems emphasize that the body can use many configurations to realize the same task outcome \citep{bernstein1967coordination}. Dynamical and constraints-based accounts focus on how interactions among the body, task, and environment organize this redundancy \citep{newell1986constraints,renshaw2010constraints}. Passive mechanical dynamics demonstrate how body and environment can generate useful movement structure \citep{mcgeer1990passive}, while central pattern generators provide a neural account of recurring motor rhythms \citep{ijspeert2008central}. Work on augmented feedback and the guidance hypothesis shows that the availability and frequency of feedback can affect practice, retention, and transfer \citep{schmidt1991frequent,winstein1994effects}. Together, these accounts motivate the idea that observations can serve different roles during practice and later action, while offering distinct explanations of the underlying mechanisms.

Recent advances in general-purpose model-based \gls*{rl} feature agents that acquire complex behaviors from interaction with little task-specific structure. In particular, \texttt{DreamerV3} has learned to play Minecraft \citep{hafner2025mastering}, perform tactile insertion \citep{palenicek2024learning}, play a real-world labyrinth game \citep{bi2024sample}, and control a quadrotor \citep{romero2025dream}. For agents acting in the physical world, these developments make the preceding distinction concrete: observations that help acquire a behavior need not remain available when the agent later executes it. Learning using privileged information formalizes this separation \citep{vapnik2009new}: extra training inputs may be exact simulator state \citep{pinto2017asymmetric} or measurements from costly \gls*{lidar} units. In \gls*{rl}, asymmetric actor--critic methods expose the critic to state unavailable to the deployed policy \citep{pinto2017asymmetric}; learning by cheating transfers behavior from a privileged teacher to a vision-based policy \citep{chen2020learning}; attention-privileged \gls*{rl} uses privileged features to guide attention during training \citep{salter2021attention}; and privileged sensing supplies training-only information to components such as critics, world models, and reward estimators \citep{hu2024privileged}. We ask whether privileged observations can support policy discovery when an agent trains directly on a physical system. We examine this question in one concrete physical setting: active flow control, which actuates a system immersed in a fluid flow to alter the flow field (a spatial map of the flow velocity) surrounding it and thereby optimize performance metrics such as the drag of the system. In our setup, the agent rotates a cylinder in flowing water to minimize or maximize the drag on it; see \cref{fig:cover}. Here, \emph{flow feedback} means supplying this estimate online as an observation, distinct from the scalar reward that specifies the task objective. We ask how privileged flow observations affect policy discovery in active flow control.

To investigate this question, we use the generalist \gls*{rl} algorithm \texttt{DreamerV3} \citep{hafner2025mastering}, without hyperparameter tuning or a task-specific inductive bias. We interface the agent with the tabletop \gls*{cwc} depicted in \cref{fig:hardware_overview} and train it directly on the physical flow to minimize or maximize the drag on a spinning cylinder. We vary the observations available during training and perform extensive ablations to assess the contribution of flow observations to policy discovery. The experimental campaign spans hundreds of hours; see \cref{appendix:experimental-accounting}.

Our active-flow-control setup has three useful properties. First, the Navier--Stokes equations govern its infinite-dimensional nonlinear dynamics. High-fidelity simulations are computationally demanding and remain sensitive to modeling choices. Training directly on the physical flow avoids selecting a simulation model and includes boundary effects and other apparatus-specific dynamics. Second, we can state the objective---drag minimization or maximization---simply through a dense reward, but the dynamics do not make effective actions obvious. Third, decades-old experimental studies provide recipes for simple, high-performance, periodic open-loop policies; see \nameref{sec:experiments:open-loop}. The setup therefore separates the existence of effective action sequences from the difficulty of discovering them.

Active flow control has a decades-long experimental and computational history: researchers have used prescribed rotation to reshape cylinder wakes and forces since the 1980s \citep{coutanceau1985influence,tokumaru1991rotary,tokumaru1993lift}. Subsequent computational feedback studies used wall-pressure or wall-shear information \citep{lee1998suboptimal}, neural networks \citep{lee1997application}, and reduced-order models \citep{mathelin2010closed}; physical wake experiments also demonstrated closed-loop control \citep{siegel2004feedback,henning2005drag}. More recently, researchers have used \gls*{rl} to discover control policies, predominantly in simulation \citep{rabault2019artificial,paris2021robust,chatzimanolakis2024drag,suarez2025flow,ye2025model}. In a physical experiment, \citet{fan2020reinforcement} trained an agent using force measurements from repeated towing runs. Earlier physical controllers compressed real-time \gls*{piv} fields to an instantaneous recirculation area for genetic-programming control \citep{gautier2015closed}, to a wake-barycenter coordinate for a prescribed opposition law \citep{varon2019adaptive}, or used sparse local velocity components for proportional surface actuation \citep{mccormick2024reactive}. Our \gls*{rl} agent instead receives the dense flow field estimate during direct physical training; we isolate how this information affects policy discovery.

\enlargethispage{\baselineskip}
\begin{mdframed}[hidealllines=true,backgroundcolor=blue!5]
\vspace{-.25cm}
\paragraph{Contributions.}
We use a \gls*{rl} agent to study how observations shape policy discovery in a real, chaotic fluid system. Our contributions are:
\begin{enumerate}[leftmargin=*, itemsep=-2pt, topsep=3pt]
\item We design and release a low-cost tabletop water channel
for active flow control experiments. The channel is built from off-the-shelf components and 3D-printed parts, and it is instrumented with a low-cost \gls*{piv} system that provides real-time full-field flow measurements; see \cref{fig:hardware_overview}.

\item We train a generalist \gls*{rl} agent directly on the chaotic physical wake rather than in simulation. To our knowledge, this is the first physical demonstration of such an agent rapidly and reliably discovering high-performing active-flow-control policies while receiving dense flow observations within minutes of real-world interaction, providing evidence that dense flow observations are a promising tool for training active-flow-control policies and, more broadly, that privileged information
can be decisive for policy discovery when training directly on a physical system.

\item Changing only the sign of the drag objective changes whether flow observations are decisive for policy discovery. In fact, we can replay action trajectories that we record during online policy execution without online observations while retaining comparable mean performance under matched, stabilized conditions. Nonetheless, when we withhold flow observations, the agent discovers high-performing drag-minimizing policies, yet fails to synthesize a high-performing drag-maximizing policy.

\item Via extensive ablations, we provide evidence that this asymmetry is related to the prevalent initial drag decrease following the actions of the agent; this physical response favors minimization but initially opposes maximization.
\end{enumerate}
The experiments provide a physical demonstration that privileged observations can be decisive for policy discovery even in the extreme scenario when we can subsequently replay the resulting action trajectories in open loop.
Practically, our findings suggest that rich observations accessible during laboratory training can be powerful even when their continuous acquisition and processing would be impractical at deployment.
\end{mdframed}

\PFFFigure{3}
\section{Results}
\label{sec:results}
When we report a drag change as a percentage without naming another reference, we normalize it by the common no-control reference defined in \nameref{sec:system:water-channel}.

\subsection{Learning active flow control}
\label{sec:fluids}
As early as the 1980s, \citet{diaz1983vortex} (among others) observed that steady cylinder rotation can reduce or suppress vortex shedding behind a circular cylinder $\mathcal{C}$ of radius $R$ in a uniform flow \citep{aljure2015influence}.
Time-dependent rotary oscillation can instead reshape the wake and alter drag and mixing \citep{tokumaru1991rotary}.
We denote the commanded angular velocity by $\omega(t)$.

For a Newtonian fluid of density $\rho$ and kinematic viscosity $\nu$, the incompressible Navier--Stokes equations govern the velocity field $\vec{u}(t,\vec{x})$ and pressure $p(t,\vec{x})$ (e.g., \citealp{acheson1990elementary}),
$
\partial_t \vec{u} + (\vec{u}\cdot\nabla)\vec{u} = -\frac{1}{\rho}\nabla p + \nu \nabla^2 \vec{u},
$ $
\nabla\cdot\vec{u} = 0,
$
subject to appropriate boundary conditions (e.g., incoming flow stream, no-slip at the walls). Commanding a rotation rate $\omega(t)$ imposes
$
    \vec{u}(t, \vec{x}) = R\omega(t)\vec{t}_{\mathcal{C}}(\vec{x})
$
 {for all} 
 $
    \vec{x} \in \partial \mathcal{C},
$
where $\vec{t}_{\mathcal{C}}(\vec{x})$ denotes the unit tangent to $\mathcal{C}$ at $\vec{x}$. The corresponding Newtonian Cauchy stress is
\begin{equation}
\label{eq:cauchy-stress}
\boldsymbol{\sigma}(t,\vec{x})
=-p(t,\vec{x})\mathbf{I}
+\rho\nu\left[
\nabla\vec{u}(t,\vec{x})
+\left(\nabla\vec{u}(t,\vec{x})\right)^{\mathsf{T}}
\right].
\end{equation}
Let $\vec{n}_{\mathcal{C}}(\vec{x})$ denote the unit normal pointing from the cylinder into the fluid. The traction $\boldsymbol{\sigma}(t,\vec{x})\vec{n}_{\mathcal{C}}(\vec{x})$ exerted by the fluid on the cylinder contains both pressure and viscous contributions. Integrating this traction over $\partial\mathcal{C}$ gives the hydrodynamic force; its component along the incoming-flow direction is drag. Controlling the rotation rate $\omega(t)$ alters the velocity and pressure fields and therefore both contributions to drag.
The geometry is illustrated in \cref{fig:cylinder-schematic}.

\PFFFigure[.32\linewidth]{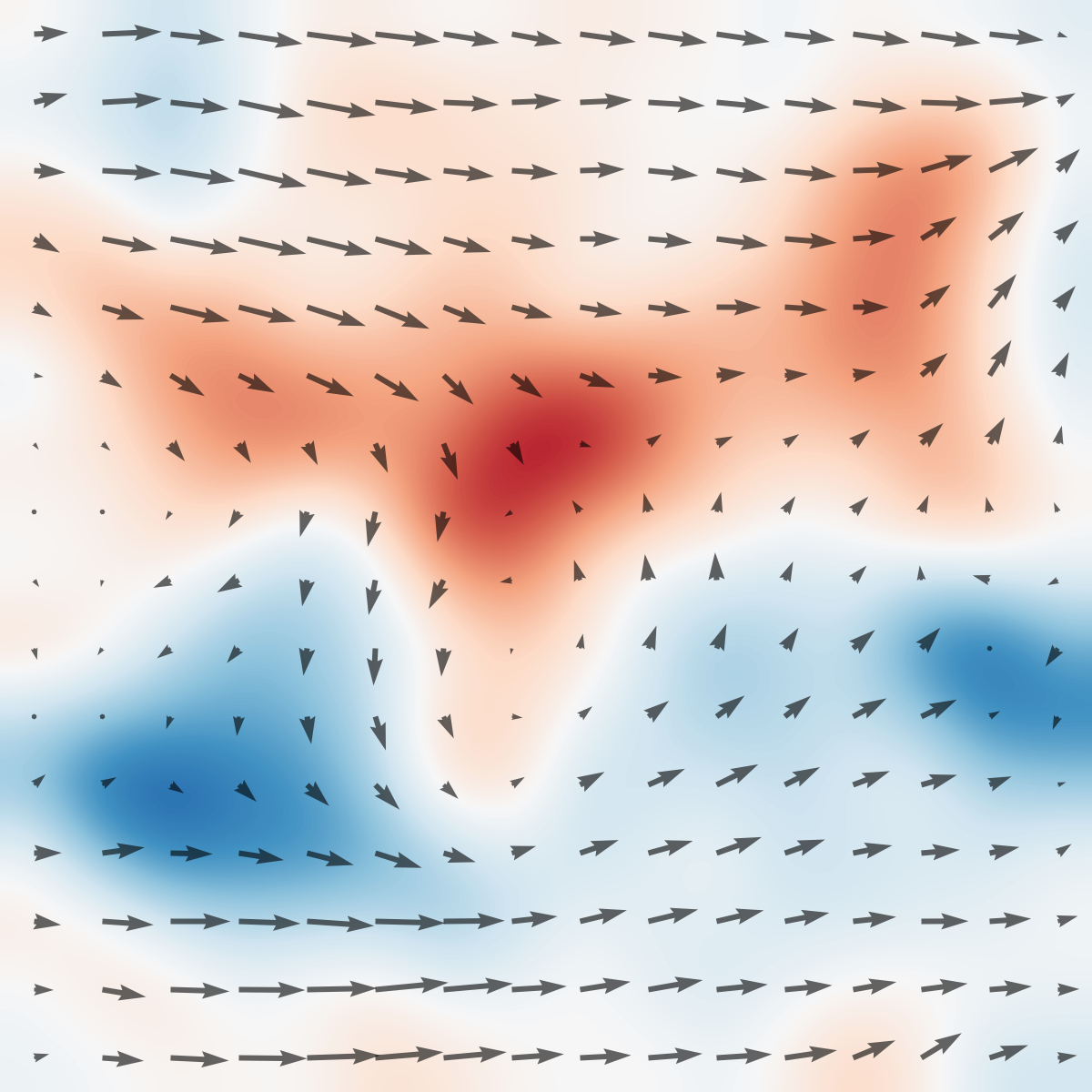}

We therefore use this drag as the performance metric that the agent minimizes or maximizes.
Despite the complexity of the system dynamics, the system admits high-performing open-loop inputs; \nameref{sec:experiments:open-loop} identifies sinusoidal examples of the form $\omega(t)=A\sin(2\pi ft)$ for $f>0$.
High-fidelity simulation of this setup is computationally demanding and sensitive to modeling and boundary-condition choices, while systematic three-dimensional studies with active control require substantial resources. We therefore built a low-cost physical platform. \Cref{fig:hardware_overview} shows the platform, and \nameref{sec:system:water-channel} describes it in detail. To produce flow observations, we developed a hardware-accelerated flow estimation pipeline that runs in real time \citep{terpin2026flow}. We tuned the pipeline using \emph{SynthPix}, a synthetic \gls*{piv} image generator that reproduces the optical setup and seeding conditions \citep{terpin2026synthpix}. We supply the resulting dense flow field estimates to the \gls*{rl} agent while training directly on the physical system.

Deploying \gls*{rl} for active flow control.
\label{sec:system:problem}
For an overview of the system, see \cref{fig:system}. We use the \texttt{DreamerV3} architecture with default hyperparameters and the standard model with $\PFFProtocolStandardModelParametersMillions$ million parameters, and report learning-algorithm ablations in \cref{appendix:rl}. The agent selects a scalar continuous action in $[-\bar A,\bar A]$ commanding the cylinder's instantaneous rotation rate, which we normalize to $[\PFFProtocolActionNormalizedLowerBound,\PFFProtocolActionNormalizedUpperBound]$ for learning. We either supply flow observations to the agent or withhold them while keeping all other observations fixed. The reduced set contains the drag measurement, the previous commanded rotation rate, and motor rotation-rate feedback; the augmented set additionally contains the dense flow field estimate produced by this pipeline.
Let $d_t$ denote the filtered drag measurement at time $t$, let $d_0$ denote the episode-start value that we measure before actuation, and let $s=+1$ for maximization and $s=-1$ for minimization. We use $r_t=s(d_t-d_0)$, so the reward measures drag change from the initial unactuated state. The drag measurement belongs to both observation sets.
The common no-control reference that we use for percentage reporting is distinct from the episode-start reward-centering value $d_0$; see \nameref{sec:discussion}.
Episodes last $\PFFProtocolEpisodeDurationS\,\mathrm{s}$. Before each episode, we wait $\PFFProtocolStabilizationS\,\mathrm{s}$ for the flow to stabilize and obtain similar, though not identical, initial conditions. Training uses a $\PFFProtocolTrainingBudgetMinutes$-minute budget ($\PFFProtocolTrainingEpisodes$ episodes), corresponding to $\PFFProtocolTrainingStepsDisplay$ environment steps.
\PFFFigure{5}

\subsection[Supplying flow observations rapidly yields high-performance policies; withholding them impedes drag-maximization training]{Supplying flow observations rapidly yields high-performance policies; withholding them impedes drag-maximization training}
\label{sec:experiments:closed-loop}

We supply or withhold flow observations while training each agent and measure the change in drag relative to its value at the beginning of each episode (a decrease for drag minimization, an increase for drag maximization). We repeat each training run \PFFProtocolMainRepetitions{} times. We perform several observation-set ablations in \cref{sec:ablations:observations-sets}.

\Cref{fig:training-curves} shows the training curves when we supply or withhold flow observations.
With flow observations, the agent learns high-performing closed-loop policies for both tasks within minutes of physical interaction; this conclusion is unchanged across the attainment thresholds examined in \cref{tab:headline-training-threshold-sensitivity}. Over the final \PFFProtocolEndpointEpisodeCount{} episodes, these policies increase drag by $\PFFPaperHeadlineMaxFlowEndpointPercentPM\%$ and reduce it by $\PFFPaperHeadlineMinFlowEndpointPercentPM\%$, respectively. However, when we withhold flow observations during training, the agent still discovers high-performing drag-minimizing policies ($\PFFPaperHeadlineMinNoFlowEndpointPercentPM\%$ drag reduction), albeit more slowly and with greater variability. No run learns a high-performing drag-maximizing policy ($\PFFPaperHeadlineMaxNoFlowEndpointPercentPM\%$ drag increase); see \nameref{appendix:headline-training-metrics} and \nameref{sec:statistical-summaries}.

The two tasks share the same physical system, admit high-performing action sequences of similar complexity, and differ in reward only by a sign. Nevertheless, flow observations are decisive for discovering the drag-maximizing behavior and much less important for drag minimization.

We investigate this asymmetry next.

\PFFFigure{6}

\subsection[Replaying action trajectories under matched conditions preserves mean performance without online observations]{Replaying action trajectories under matched conditions preserves mean performance without online observations}
\label{sec:results:feedforward}

Throughout training, we record action trajectories from the agent receiving the augmented observation set, including flow observations. We replay each trajectory \PFFProtocolReplayRepeats{} times without online observations and measure the resulting performance. Between trials, we use the reset procedure in \nameref{sec:fluids}.

\Cref{fig:results:feedforward:lc} compares online policy execution with open-loop replay of the recorded action trajectories. Under matched, stabilized conditions, open-loop replay attains episode-average filtered-drag changes similar to those from online execution, including when we replay trajectories from early in training. Averaged over the final \PFFProtocolEndpointEpisodeCount{} checkpoints within each training repetition, replay increases drag by $\PFFPaperReplayMaxEndpointPercentPM\%$ in drag maximization and reduces it by $\PFFPaperReplayMinEndpointPercentPM\%$ in drag minimization, relative to the no-control mean; see \cref{appendix:temporal-structure} for the paired comparison across all checkpoints. Thus, under this protocol, replay retains comparable mean performance in both tasks, with a modest advantage for online execution in drag maximization.

Flow observations substantially improve discovery of a high-performing drag-maximizing policy, while subsequent open-loop replay of action trajectories from online execution attains comparable mean performance under matched, stabilized conditions. The agent can therefore benefit from richer information during policy discovery than open-loop replay requires to reproduce the recorded behavior.

\PFFFigure{7}

Comparable mean performance under matched, stabilized conditions does not show that online feedback plays no role during execution. In \cref{appendix:temporal-structure}, we therefore probe whether its contribution appears in the within-episode dynamics by comparing online--replay differences in the measured motor-rotation-rate and filtered-drag spectra with variation among repeated replays.

\subsection[A prevalent negative drag transient]{A prevalent negative drag transient}
\label{sec:theory}
When we withhold flow observations, the agent nevertheless discovers a drag-minimizing policy. The drag-maximization discovery failure without flow observations recurs across the observation-set and learning-algorithm ablations in \cref{sec:ablations:observations-sets,appendix:rl}.
An initial material drag decrease occurs in \PFFPaperNegativeOnlineMaterialPercent\% of the online \texttt{DreamerV3} trajectories that we analyze. This material decrease is observed also in \PFFPaperNegativeOnlineSustainedPositivePrecededByMaterialNegativePercent\% of the trajectories that attain a sustained drag increase; see \cref{fig:results:feedforward:lc,appendix:negative-response-analysis} for a detailed sensitivity analysis. We call this ordered decrease-then-increase response a trajectory-level inverse response and we attribute the observed learning asymmetry to the fact that initial decrease immediately favors minimization but opposes maximization.
At a related Reynolds number, \citet{aljure2015influence} report rotation-induced drag reduction and accompanying changes in separation and wake topology, providing qualitative precedent for the initial decrease. Unsteady, rotation-dependent force and wake evolution offer further context \citep{badr1990unsteady}. Neither study establishes the temporal ordering that we measure or the mechanism in the confined apparatus.
Similar inverse transients arise in non-minimum-phase input--output models of bicycle steering \citep{aastrom2005bicycle}, walking \citep{dallali2009using,wee2013design}, and vertical- and conventional-takeoff-and-landing aircraft \citep{hauser1992nonlinear,tomlin1995output}.
Recent work evaluates learning-based controllers on non-minimum-phase benchmarks \citep{agyei2026deep} and develops specialized control methods for non-minimum-phase dynamics \citep{charfeddine2025adaptive,van2018inversion,zhou2018inversion}; see also the \nameref{sec:theory-excess-cost}.

Flow observations repeatedly enable the agent to discover a high-performing drag-maximizing policy despite the initially adverse response.

\FloatBarrier
\section{Discussion}
\label{sec:discussion}
\label{sec:theory-excess-cost}
\label{sec:privileged-info}
\label{sec:observation-transition}
\label{sec:discussion-limitations}
These experiments separate two questions often conflated: which observations help an agent discover high-performing behavior and which observations an online policy needs during execution. Replay replaces the policy with a fixed command sequence and therefore tests if the online corrections are needed to retain online performance.

Classical control theory has primarily examined how observations constrain robust execution. \citet{doyle1978guaranteed} showed that standard linear--quadratic--Gaussian regulators need not have guaranteed robustness margins. In stick balancing, fixation location, sensory delay, and noise affect robustness \citep{leong2016understanding}; related observation and perception limits affect the sample efficiency and performance of learned perception-based controllers \citep{xu2021learned}.
Our experiments raise the complementary question of how the observation set affects the reliability of policy discovery. Both drag objectives admit high-performing sinusoidal action sequences. Nevertheless, under the tested reward specification and $\PFFProtocolTrainingBudgetMinutes$-minute training budget, withholding flow observations sharply impedes drag-maximization training, whereas drag-minimization training still succeeds, albeit more slowly and less reliably. The existence of an effective action sequence therefore does not imply that the tested training procedure will reliably discover it from a restricted observation set.
The observed asymmetry arises jointly from the flow dynamics, available observations, and reward specification. In particular, the prevalent inverse response observed in the online trajectories of the \gls*{rl} agent makes the immediately available drag signal initially adverse to maximization, identifying a key difficulty in the drag-maximization setting. Replacing the episode-start reward-centering value $d_0$ by a fixed value $b$ gives $s(d_t-b)=r_t+s(d_0-b)$, where $s$ is the task sign. For a fixed initial state and episode horizon, the added return is independent of the action trajectory. With a fixed, policy-independent initial-state distribution, this recentering preserves the expected ordering of policies. It also leaves the physical drag response unchanged for a fixed action trajectory. Numerical recentering or policy-invariant shaping can nevertheless alter finite-sample training under function approximation through exploration and temporal credit assignment.
\citet{lee2024active} analyze active data collection for control-oriented identification of nonlinear systems and relate excess control cost to both the sensitivity of control performance to estimation error and the difficulty of identifying the dynamics under exploration. Their analysis does not apply directly to policy discovery in our physical \gls*{rl} setting. Bridging the two settings therefore presents two opportunities: developing a model of the observation-dependent discovery asymmetry that we measure here and using that model to design targeted experiments that test whether flow observations supply information along directions relevant to the sustained drag response.

Replay provides a constructive feasibility test of whether, once we identify an effective action trajectory, open-loop execution can reproduce high-performing behavior without flow observations. When we apply action trajectories from online policy execution as fixed sequences under matched, stabilized conditions, they retain comparable mean performance. This result demonstrates that the task admits an effective observation-free action sequence that the \gls*{rl} algorithm can find under the tested conditions; the difficulty of drag-maximization training with restricted observations therefore lies in reliably discovering a successful behavior. Richer observations can thus be valuable for discovery even when reproducing a recorded action trajectory requires no observations under matched conditions. Online execution retains a modest advantage in drag maximization. \Cref{appendix:temporal-structure} reports descriptive paired intervals that exclude zero for the filtered-drag response in both tasks but include zero for the measured motor-rotation-rate response. The filtered-drag differences provide targets for the robustness and observation-ablation experiments that we propose below.

Researchers in both supervised learning and \gls*{rl} have recognized the relevance of different observations at training and deployment. In supervised learning, \citet{vapnik2009new} formalized \emph{learning using privileged information}, where additional inputs are available only at training time. In \gls*{rl}, a critic can receive simulator state while the actor operates from images at deployment \citep{pinto2017asymmetric}. Privileged teachers or state-privileged companion agents can instead train policies that ultimately operate from visual observations alone \citep{chen2020learning,salter2021attention}.
\citet{hu2024privileged} similarly use privileged sensing in critics, world models, reward estimators, and other training-only components to improve a target policy with a fixed deployment observation set.
Event-triggered \gls*{rl} learns control and communication behavior jointly to reduce sampling \citep{baumann2018deep}. In action-contingent observability, an agent can pay to reveal the current state \citep{nam2021reinforcement}; in continuous-time control, irregular observation schedules can outperform regular sampling when observations are costly \citep{holt2023active}. Minimum-attention control similarly seeks control laws that meet performance requirements while limiting their dependence on state and time \citep{brockett1997minimum}.
Together with our results, the adaptive-sensing literature and the human motor-learning literature motivate a broader research program on the dynamic use of observations: when should an agent use each observation during policy discovery, online execution, or recovery after conditions change? The answer could reveal how information use changes with practice, guide sensor selection, and indicate when an agent should activate or query each available sensor.

The present study has several limitations, each of which defines an exciting research direction. First, we report quantitative results only for the tested reward specification, training budgets, observation sets, and \hyperref[sec:uncertainties]{strongly wall-confined wake}. This well-defined scope provides a starting point for mapping the discovery asymmetry across reward formulations and flow regimes. Experiments that vary reward recentering or policy-invariant shaping would test the sensitivity of policy discovery to the reward specification. Repeating the comparison across blockage ratios, inflow speeds, and body geometries would establish how broadly the asymmetry transfers.

Second, the constructive replay result establishes observation-free feasibility under matched, stabilized operating conditions; we have not yet tested disturbance rejection or transfer to a different flow regime. The controlled protocol provides a baseline for experiments that systematically perturb the initial wake, action timing, or inflow. Together, these experiments would determine when feedback contributes to execution and robustness beyond the feasibility established by replay.

Third, the experiments do not identify how the use of flow observations changes during policy discovery or which component of training benefits from them. Replayability is consistent with several possibilities: the policy may become less sensitive to particular measured inputs, temporal structure may arise from the recurrent dynamics of \texttt{DreamerV3} \citep{hafner2025mastering}, or repeatability may emerge from the coupled policy--flow dynamics. Tracking sensitivity to individual observation channels over training and inspecting recurrent-state, representation-level, and prediction-level diagnostics could distinguish these possibilities. The same analyses could test whether flow observations alter the predictive representations or losses of the learned world model and whether any such change benefits policy discovery.

The comparison with human motor learning has a corresponding scope: it is methodological rather than mechanistic; the present experiments contain no human data; and fixed-trajectory replay is not equivalent to a retention or transfer test. In a novel visuomotor task, \citet{sailer2005eye} found that gaze changed from following a poorly controlled cursor during exploration to anticipating desired cursor positions and, later, fixating the targets. This changing pattern of information sampling motivates asking whether the use of observations also shifts during practice. Guidance-hypothesis accounts interpret some effects of reduced augmented feedback through dependence on that feedback \citep{schmidt1991frequent,winstein1994effects}, whereas perception--action and constraints-based accounts emphasize exploration and attunement to task-relevant information \citep{jacobs2007direct,renshaw2010constraints}. Pairing time-resolved measurements of information use with human retention or transfer tests would extend the comparison while leaving these competing explanations open.

The most direct systems opportunity is to make the availability of full-field \gls*{piv} a decision variable. Providing it on demand at an explicit acquisition and processing cost would test when the agent requests richer observations, when it can reduce their sampling rate, and whether restoring them recovers performance after the physical system changes. This formulation connects policy discovery to adaptive sensing and offers a route toward experimental systems that allocate sensing effort where it has the greatest value.

Taken together, the experiments show that dense flow observations can be decisive for policy discovery even when recorded action trajectories retain comparable mean performance during open-loop replay under matched, stabilized conditions. For active flow control, full-field \gls*{piv} can therefore serve not only as a diagnostic but also as a training input. The research opportunities above extend this result from asking which observations matter to asking when they matter, why they help, and how to schedule their acquisition.
Practically, our findings suggest that rich observations accessible during laboratory training can be powerful even when their continuous acquisition and processing would be impractical at deployment.

\begingroup
\small
\paragraph{Data availability.}
The experimental data, water-channel designs, and fabrication files supporting this study are freely available at \url{https://activefluidcontrol.com}.

\paragraph{Code availability.}
The code used in this study is freely available through the individual projects linked at \url{https://activefluidcontrol.com}. The accompanying documentation provides software versions, experimental and analysis parameters, and instructions for reproducing the results.

\paragraph{Acknowledgements.}
We thank F. Noca and R. Bernhard for input on the fluid-dynamics hardware; A. K. Marcus, M. M. Felix and D. Wagner for help with electronics and fabrication; and B. Lee, T. Anastasios, C. Sferrazza, P. Abbeel, J. Lygeros, N. Lanzetti, S. Bolognani, P. Grontas and F. Blanchini for scientific discussions.

\paragraph{Funding.}
This work was supported by an ETH RobotX research grant funded through the ETH Zurich Foundation. The funders had no role in study design, data collection and analysis, the decision to publish, or preparation of the manuscript.

\paragraph{Author contributions.}
Conceptualization, Methodology: A.T., R.D. Software, Hardware, Investigation, Formal Analysis, Data Curation, Writing--Original Draft: A.T. Supervision: R.D. Writing--Review: A.T., R.D.
\paragraph{Competing interests.}
The authors declare no competing financial or non-financial interests.

\par\endgroup

\clearpage
\phantomsection
\addcontentsline{toc}{section}{References}
\bibliographystyle{sn-nature}
\bibliography{main}

\begin{thebibliography}{10}
\expandafter\ifx\csname url\endcsname\relax
  \def\url#1{\burl{#1}}\fi
\expandafter\ifx\csname urlprefix\endcsname\relax\def\urlprefix{URL }\fi
\providecommand{\bibinfo}[2]{#2}
\providecommand{\eprint}[2][]{\url{#2}}
\providecommand{\doi}[1]{\url{https://doi.org/#1}}
\bibcommenthead

\bibitem[Lee and Aronson(1974)]{lee1974visual}
\bibinfo{author}{Lee, D.~N.} \& \bibinfo{author}{Aronson, E.}
\newblock \bibinfo{title}{Visual proprioceptive control of standing in human
  infants}.
\newblock \emph{\bibinfo{journal}{Perception \& Psychophysics}}
  \textbf{\bibinfo{volume}{15}}, \bibinfo{pages}{529--532}
  (\bibinfo{year}{1974}).

\bibitem[Anderson et~al.(2019)]{anderson2019balance}
\bibinfo{author}{Anderson, D.~I.}, \bibinfo{author}{He, M.},
  \bibinfo{author}{Gutierrez, P.}, \bibinfo{author}{Uchiyama, I.} \&
  \bibinfo{author}{Campos, J.~J.}
\newblock \bibinfo{title}{Do balance demands induce shifts in visual
  proprioception in crawling infants?}
\newblock \emph{\bibinfo{journal}{Frontiers in Psychology}}
  \textbf{\bibinfo{volume}{10}}, \bibinfo{pages}{1388} (\bibinfo{year}{2019}).

\bibitem[Iosa et~al.(2012)]{iosa2012walking}
\bibinfo{author}{Iosa, M.}, \bibinfo{author}{Fusco, A.},
  \bibinfo{author}{Morone, G.} \& \bibinfo{author}{Paolucci, S.}
\newblock \bibinfo{title}{Walking there: environmental influence on
  walking-distance estimation}.
\newblock \emph{\bibinfo{journal}{Behavioural Brain Research}}
  \textbf{\bibinfo{volume}{226}}, \bibinfo{pages}{124--132}
  (\bibinfo{year}{2012}).

\bibitem[Souman et~al.(2009)]{souman2009walking}
\bibinfo{author}{Souman, J.~L.}, \bibinfo{author}{Frissen, I.},
  \bibinfo{author}{Sreenivasa, M.~N.} \& \bibinfo{author}{Ernst, M.~O.}
\newblock \bibinfo{title}{Walking straight into circles}.
\newblock \emph{\bibinfo{journal}{Current Biology}}
  \textbf{\bibinfo{volume}{19}}, \bibinfo{pages}{1538--1542}
  (\bibinfo{year}{2009}).

\bibitem[Naaman et~al.(2023)]{naaman2023young}
\bibinfo{author}{Naaman, T.}, \bibinfo{author}{Hayek, R.},
  \bibinfo{author}{Gutman, I.} \& \bibinfo{author}{Springer, S.}
\newblock \bibinfo{title}{Young, but not in the dark—the influence of reduced
  lighting on gait stability in middle-aged adults}.
\newblock \emph{\bibinfo{journal}{PLOS ONE}} \textbf{\bibinfo{volume}{18}},
  \bibinfo{pages}{e0280535} (\bibinfo{year}{2023}).

\bibitem[Adams(1971)]{adams1971closed}
\bibinfo{author}{Adams, J.~A.}
\newblock \bibinfo{title}{A closed-loop theory of motor learning}.
\newblock \emph{\bibinfo{journal}{Journal of Motor Behavior}}
  \textbf{\bibinfo{volume}{3}}, \bibinfo{pages}{111--150}
  (\bibinfo{year}{1971}).

\bibitem[Schmidt(1975)]{schmidt1975schema}
\bibinfo{author}{Schmidt, R.~A.}
\newblock \bibinfo{title}{A schema theory of discrete motor skill learning}.
\newblock \emph{\bibinfo{journal}{Psychological Review}}
  \textbf{\bibinfo{volume}{82}}, \bibinfo{pages}{225--260}
  (\bibinfo{year}{1975}).

\bibitem[van Rossum(1990)]{vanrossum1990schema}
\bibinfo{author}{van Rossum, J. H.~A.}
\newblock \bibinfo{title}{Schmidt's schema theory: The empirical base of the
  variability of practice hypothesis: A critical analysis}.
\newblock \emph{\bibinfo{journal}{Human Movement Science}}
  \textbf{\bibinfo{volume}{9}}, \bibinfo{pages}{387--435}
  (\bibinfo{year}{1990}).

\bibitem[Jacobs and Michaels(2007)]{jacobs2007direct}
\bibinfo{author}{Jacobs, D.~M.} \& \bibinfo{author}{Michaels, C.~F.}
\newblock \bibinfo{title}{Direct learning}.
\newblock \emph{\bibinfo{journal}{Ecological Psychology}}
  \textbf{\bibinfo{volume}{19}}, \bibinfo{pages}{321--349}
  (\bibinfo{year}{2007}).

\bibitem[Bernstein(1967)]{bernstein1967coordination}
\bibinfo{author}{Bernstein, N.~A.}
\newblock \emph{\bibinfo{title}{The Co-ordination and Regulation of Movements}}
   (\bibinfo{publisher}{Pergamon Press}, \bibinfo{address}{Oxford},
  \bibinfo{year}{1967}).

\bibitem[Newell(1986)]{newell1986constraints}
\bibinfo{author}{Newell, K.~M.}
\newblock \bibinfo{title}{ in \textit{Constraints on the development of
  coordination}} \emph{\bibinfo{booktitle}{Motor Development in Children:
  Aspects of Coordination and Control}} \bibinfo{pages}{341--360}
  (\bibinfo{publisher}{Martinus Nijhoff}, \bibinfo{address}{Dordrecht},
  \bibinfo{year}{1986}).

\bibitem[Renshaw et~al.(2010)]{renshaw2010constraints}
\bibinfo{author}{Renshaw, I.}, \bibinfo{author}{Chow, J.~Y.},
  \bibinfo{author}{Davids, K.} \& \bibinfo{author}{Hammond, J.}
\newblock \bibinfo{title}{A constraints-led perspective to understanding skill
  acquisition and game play: A basis for integration of motor learning theory
  and physical education praxis?}
\newblock \emph{\bibinfo{journal}{Physical Education and Sport Pedagogy}}
  \textbf{\bibinfo{volume}{15}}, \bibinfo{pages}{117--137}
  (\bibinfo{year}{2010}).

\bibitem[McGeer(1990)]{mcgeer1990passive}
\bibinfo{author}{McGeer, T.}
\newblock \bibinfo{title}{Passive dynamic walking}.
\newblock \emph{\bibinfo{journal}{The International Journal of Robotics
  Research}} \textbf{\bibinfo{volume}{9}}, \bibinfo{pages}{62--82}
  (\bibinfo{year}{1990}).

\bibitem[Ijspeert(2008)]{ijspeert2008central}
\bibinfo{author}{Ijspeert, A.~J.}
\newblock \bibinfo{title}{Central pattern generators for locomotion control in
  animals and robots: a review}.
\newblock \emph{\bibinfo{journal}{Neural Networks}}
  \textbf{\bibinfo{volume}{21}}, \bibinfo{pages}{642--653}
  (\bibinfo{year}{2008}).

\bibitem[Schmidt(1991)]{schmidt1991frequent}
\bibinfo{author}{Schmidt, R.~A.}
\newblock \bibinfo{title}{ in \textit{Frequent augmented feedback can degrade
  learning: Evidence and interpretations}} \emph{\bibinfo{booktitle}{Tutorials
  in motor neuroscience}}, Vol.~\bibinfo{volume}{62} of
  \emph{\bibinfo{series}{NATO ASI Series}} \bibinfo{pages}{59--75}
  (\bibinfo{publisher}{Springer, Dordrecht}, \bibinfo{year}{1991}).

\bibitem[Winstein et~al.(1994)]{winstein1994effects}
\bibinfo{author}{Winstein, C.~J.}, \bibinfo{author}{Pohl, P.~S.} \&
  \bibinfo{author}{Lewthwaite, R.}
\newblock \bibinfo{title}{Effects of physical guidance and knowledge of results
  on motor learning: support for the guidance hypothesis}.
\newblock \emph{\bibinfo{journal}{Research Quarterly for Exercise and Sport}}
  \textbf{\bibinfo{volume}{65}}, \bibinfo{pages}{316--323}
  (\bibinfo{year}{1994}).

\bibitem[Hafner et~al.(2025)]{hafner2025mastering}
\bibinfo{author}{Hafner, D.}, \bibinfo{author}{Pasukonis, J.},
  \bibinfo{author}{Ba, J.} \& \bibinfo{author}{Lillicrap, T.}
\newblock \bibinfo{title}{Mastering diverse control tasks through world
  models}.
\newblock \emph{\bibinfo{journal}{Nature}} \textbf{\bibinfo{volume}{640}},
  \bibinfo{pages}{647--653} (\bibinfo{year}{2025}).

\bibitem[Palenicek et~al.(2024)]{palenicek2024learning}
\bibinfo{author}{Palenicek, D.}, \bibinfo{author}{Gruner, T.},
  \bibinfo{author}{Schneider, T.}, \bibinfo{author}{B{\"o}hm, A.},
  \bibinfo{author}{Lenz, J.}, \bibinfo{author}{Pfenning, I.},
  \bibinfo{author}{Kr{\"a}mer, E.} \& \bibinfo{author}{Peters, J.}
  \emph{\bibinfo{title}{Learning tactile insertion in the real world}}.
\newblock \emph{\bibinfo{booktitle}{40th Anniversary of the {IEEE}
  International Conference on Robotics and Automation ({ICRA@40})}}
  (\bibinfo{year}{2024}).

\bibitem[Bi and D'Andrea(2024)]{bi2024sample}
\bibinfo{author}{Bi, T.} \& \bibinfo{author}{D'Andrea, R.}
  \emph{\bibinfo{title}{Sample-efficient learning to solve a real-world
  labyrinth game using data-augmented model-based reinforcement learning}}.
\newblock \emph{\bibinfo{booktitle}{2024 {IEEE} International Conference on
  Robotics and Automation ({ICRA})}}, \bibinfo{pages}{7455--7460}
  (\bibinfo{year}{2024}).

\bibitem[Romero et~al.(2026)]{romero2025dream}
\bibinfo{author}{Romero, A.}, \bibinfo{author}{Shenai, A.},
  \bibinfo{author}{Geles, I.}, \bibinfo{author}{Aljalbout, E.} \&
  \bibinfo{author}{Scaramuzza, D.} \emph{\bibinfo{title}{Dream to fly:
  Model-based reinforcement learning for vision-based drone flight}}.
\newblock \emph{\bibinfo{booktitle}{2026 {IEEE} International Conference on
  Robotics and Automation ({ICRA})}} (\bibinfo{address}{Vienna, Austria},
  \bibinfo{year}{2026}).

\bibitem[Vapnik and Vashist(2009)]{vapnik2009new}
\bibinfo{author}{Vapnik, V.} \& \bibinfo{author}{Vashist, A.}
\newblock \bibinfo{title}{A new learning paradigm: Learning using privileged
  information}.
\newblock \emph{\bibinfo{journal}{Neural Networks}}
  \textbf{\bibinfo{volume}{22}}, \bibinfo{pages}{544--557}
  (\bibinfo{year}{2009}).

\bibitem[Pinto et~al.(2018)]{pinto2017asymmetric}
\bibinfo{author}{Pinto, L.}, \bibinfo{author}{Andrychowicz, M.},
  \bibinfo{author}{Welinder, P.}, \bibinfo{author}{Zaremba, W.} \&
  \bibinfo{author}{Abbeel, P.} \emph{\bibinfo{title}{Asymmetric actor critic
  for image-based robot learning}}.
\newblock \emph{\bibinfo{booktitle}{Proceedings of Robotics: Science and
  Systems}} (\bibinfo{address}{Pittsburgh, Pennsylvania},
  \bibinfo{year}{2018}).

\bibitem[Chen et~al.(2020)]{chen2020learning}
\bibinfo{author}{Chen, D.}, \bibinfo{author}{Zhou, B.},
  \bibinfo{author}{Koltun, V.} \& \bibinfo{author}{Kr{\"a}henb{\"u}hl, P.}
  \emph{\bibinfo{title}{Learning by cheating}}.
\newblock \emph{\bibinfo{booktitle}{Proceedings of the Conference on Robot
  Learning}}, Vol. \bibinfo{volume}{100} of \emph{\bibinfo{series}{Proceedings
  of Machine Learning Research}}, \bibinfo{pages}{66--75}
  (\bibinfo{publisher}{PMLR}, \bibinfo{year}{2020}).

\bibitem[Salter et~al.(2021)]{salter2021attention}
\bibinfo{author}{Salter, S.}, \bibinfo{author}{Rao, D.},
  \bibinfo{author}{Wulfmeier, M.}, \bibinfo{author}{Hadsell, R.} \&
  \bibinfo{author}{Posner, I.} \emph{\bibinfo{title}{Attention-privileged
  reinforcement learning}}.
\newblock \emph{\bibinfo{booktitle}{Proceedings of the 2020 Conference on Robot
  Learning}}, Vol. \bibinfo{volume}{155} of \emph{\bibinfo{series}{Proceedings
  of Machine Learning Research}}, \bibinfo{pages}{394--408}
  (\bibinfo{publisher}{PMLR}, \bibinfo{year}{2021}).

\bibitem[Hu et~al.(2024)]{hu2024privileged}
\bibinfo{author}{Hu, E.~S.}, \bibinfo{author}{Springer, J.},
  \bibinfo{author}{Rybkin, O.} \& \bibinfo{author}{Jayaraman, D.}
  \emph{\bibinfo{title}{Privileged sensing scaffolds reinforcement learning}}.
\newblock \emph{\bibinfo{booktitle}{The Twelfth International Conference on
  Learning Representations}} (\bibinfo{year}{2024}).

\bibitem[Coutanceau and M{\'e}nard(1985)]{coutanceau1985influence}
\bibinfo{author}{Coutanceau, M.} \& \bibinfo{author}{M{\'e}nard, C.}
\newblock \bibinfo{title}{Influence of rotation on the near-wake development
  behind an impulsively started circular cylinder}.
\newblock \emph{\bibinfo{journal}{Journal of Fluid Mechanics}}
  \textbf{\bibinfo{volume}{158}}, \bibinfo{pages}{399--446}
  (\bibinfo{year}{1985}).

\bibitem[Tokumaru and Dimotakis(1991)]{tokumaru1991rotary}
\bibinfo{author}{Tokumaru, P.~T.} \& \bibinfo{author}{Dimotakis, P.~E.}
\newblock \bibinfo{title}{Rotary oscillation control of a cylinder wake}.
\newblock \emph{\bibinfo{journal}{Journal of Fluid Mechanics}}
  \textbf{\bibinfo{volume}{224}}, \bibinfo{pages}{77--90}
  (\bibinfo{year}{1991}).

\bibitem[Tokumaru and Dimotakis(1993)]{tokumaru1993lift}
\bibinfo{author}{Tokumaru, P.~T.} \& \bibinfo{author}{Dimotakis, P.~E.}
\newblock \bibinfo{title}{The lift of a cylinder executing rotary motions in a
  uniform flow}.
\newblock \emph{\bibinfo{journal}{Journal of Fluid Mechanics}}
  \textbf{\bibinfo{volume}{255}}, \bibinfo{pages}{1--10}
  (\bibinfo{year}{1993}).

\bibitem[Lee et~al.(1998)]{lee1998suboptimal}
\bibinfo{author}{Lee, C.}, \bibinfo{author}{Kim, J.} \& \bibinfo{author}{Choi,
  H.}
\newblock \bibinfo{title}{Suboptimal control of turbulent channel flow for drag
  reduction}.
\newblock \emph{\bibinfo{journal}{Journal of Fluid Mechanics}}
  \textbf{\bibinfo{volume}{358}}, \bibinfo{pages}{245--258}
  (\bibinfo{year}{1998}).

\bibitem[Lee et~al.(1997)]{lee1997application}
\bibinfo{author}{Lee, C.}, \bibinfo{author}{Kim, J.}, \bibinfo{author}{Babcock,
  D.} \& \bibinfo{author}{Goodman, R.}
\newblock \bibinfo{title}{Application of neural networks to turbulence control
  for drag reduction}.
\newblock \emph{\bibinfo{journal}{Physics of Fluids}}
  \textbf{\bibinfo{volume}{9}}, \bibinfo{pages}{1740--1747}
  (\bibinfo{year}{1997}).

\bibitem[Mathelin et~al.(2010)]{mathelin2010closed}
\bibinfo{author}{Mathelin, L.}, \bibinfo{author}{Abbas-Turki, M.},
  \bibinfo{author}{Pastur, L.} \& \bibinfo{author}{Abou-Kandil, H.}
\newblock \bibinfo{title}{Closed-loop fluid flow control using a low
  dimensional model}.
\newblock \emph{\bibinfo{journal}{Mathematical and Computer Modelling}}
  \textbf{\bibinfo{volume}{52}}, \bibinfo{pages}{1161--1168}
  (\bibinfo{year}{2010}).

\bibitem[Siegel et~al.(2004)]{siegel2004feedback}
\bibinfo{author}{Siegel, S.}, \bibinfo{author}{Cohen, K.} \&
  \bibinfo{author}{McLaughlin, T.} \emph{\bibinfo{title}{Feedback control of a
  circular cylinder wake in a water tunnel experiment}}.
\newblock \emph{\bibinfo{booktitle}{42nd {AIAA} Aerospace Sciences Meeting and
  Exhibit}} (\bibinfo{publisher}{American Institute of Aeronautics and
  Astronautics}, \bibinfo{year}{2004}).

\bibitem[Henning and King(2005)]{henning2005drag}
\bibinfo{author}{Henning, L.} \& \bibinfo{author}{King, R.}
  \emph{\bibinfo{title}{Drag reduction by closed-loop control of a separated
  flow over a bluff body with a blunt trailing edge}}.
\newblock \emph{\bibinfo{booktitle}{Proceedings of the 44th {IEEE} Conference
  on Decision and Control, and the European Control Conference 2005}},
  \bibinfo{pages}{494--499} (\bibinfo{year}{2005}).

\bibitem[Rabault et~al.(2019)]{rabault2019artificial}
\bibinfo{author}{Rabault, J.}, \bibinfo{author}{Kuchta, M.},
  \bibinfo{author}{Jensen, A.}, \bibinfo{author}{R{\'e}glade, U.} \&
  \bibinfo{author}{Cerardi, N.}
\newblock \bibinfo{title}{Artificial neural networks trained through deep
  reinforcement learning discover control strategies for active flow control}.
\newblock \emph{\bibinfo{journal}{Journal of Fluid Mechanics}}
  \textbf{\bibinfo{volume}{865}}, \bibinfo{pages}{281--302}
  (\bibinfo{year}{2019}).

\bibitem[Paris et~al.(2021)]{paris2021robust}
\bibinfo{author}{Paris, R.}, \bibinfo{author}{Beneddine, S.} \&
  \bibinfo{author}{Dandois, J.}
\newblock \bibinfo{title}{Robust flow control and optimal sensor placement
  using deep reinforcement learning}.
\newblock \emph{\bibinfo{journal}{Journal of Fluid Mechanics}}
  \textbf{\bibinfo{volume}{913}}, \bibinfo{pages}{A25} (\bibinfo{year}{2021}).

\bibitem[Chatzimanolakis et~al.(2024)]{chatzimanolakis2024drag}
\bibinfo{author}{Chatzimanolakis, M.}, \bibinfo{author}{Weber, P.} \&
  \bibinfo{author}{Koumoutsakos, P.}
\newblock \bibinfo{title}{Learning in two dimensions and controlling in three:
  Generalizable drag reduction strategies for flows past circular cylinders
  through deep reinforcement learning}.
\newblock \emph{\bibinfo{journal}{Physical Review Fluids}}
  \textbf{\bibinfo{volume}{9}}, \bibinfo{pages}{043902} (\bibinfo{year}{2024}).

\bibitem[Su{\'a}rez et~al.(2025)]{suarez2025flow}
\bibinfo{author}{Su{\'a}rez, P.}, \bibinfo{author}{Alc{\'a}ntara-{\'A}vila,
  F.}, \bibinfo{author}{Rabault, J.}, \bibinfo{author}{Mir{\'o}, A.},
  \bibinfo{author}{Font, B.}, \bibinfo{author}{Lehmkuhl, O.} \&
  \bibinfo{author}{Vinuesa, R.}
\newblock \bibinfo{title}{Flow control of three-dimensional cylinders
  transitioning to turbulence via multi-agent reinforcement learning}.
\newblock \emph{\bibinfo{journal}{Communications Engineering}}
  \textbf{\bibinfo{volume}{4}}, \bibinfo{pages}{113} (\bibinfo{year}{2025}).

\bibitem[Ye and Elsheikh(2025)]{ye2025model}
\bibinfo{author}{Ye, M.} \& \bibinfo{author}{Elsheikh, A.~H.}
\newblock \bibinfo{title}{Model-based reinforcement learning for active flow
  control}.
\newblock \emph{\bibinfo{journal}{Physics of Fluids}}
  \textbf{\bibinfo{volume}{37}}, \bibinfo{pages}{093363}
  (\bibinfo{year}{2025}).

\bibitem[Fan et~al.(2020)]{fan2020reinforcement}
\bibinfo{author}{Fan, D.}, \bibinfo{author}{Yang, L.}, \bibinfo{author}{Wang,
  Z.}, \bibinfo{author}{Triantafyllou, M.~S.} \& \bibinfo{author}{Karniadakis,
  G.~E.}
\newblock \bibinfo{title}{Reinforcement learning for bluff body active flow
  control in experiments and simulations}.
\newblock \emph{\bibinfo{journal}{Proceedings of the National Academy of
  Sciences}} \textbf{\bibinfo{volume}{117}}, \bibinfo{pages}{26091--26098}
  (\bibinfo{year}{2020}).

\bibitem[Gautier et~al.(2015)]{gautier2015closed}
\bibinfo{author}{Gautier, N.}, \bibinfo{author}{Aider, J.-L.},
  \bibinfo{author}{Duriez, T.}, \bibinfo{author}{Noack, B.~R.},
  \bibinfo{author}{Segond, M.} \& \bibinfo{author}{Abel, M.}
\newblock \bibinfo{title}{Closed-loop separation control using machine
  learning}.
\newblock \emph{\bibinfo{journal}{Journal of Fluid Mechanics}}
  \textbf{\bibinfo{volume}{770}}, \bibinfo{pages}{442--457}
  (\bibinfo{year}{2015}).

\bibitem[Varon et~al.(2019)]{varon2019adaptive}
\bibinfo{author}{Varon, E.}, \bibinfo{author}{Aider, J.-L.},
  \bibinfo{author}{Eulalie, Y.}, \bibinfo{author}{Edwige, S.} \&
  \bibinfo{author}{Gilotte, P.}
\newblock \bibinfo{title}{Adaptive control of the dynamics of a fully turbulent
  bimodal wake using real-time {PIV}}.
\newblock \emph{\bibinfo{journal}{Experiments in Fluids}}
  \textbf{\bibinfo{volume}{60}}, \bibinfo{pages}{124} (\bibinfo{year}{2019}).

\bibitem[McCormick et~al.(2024)]{mccormick2024reactive}
\bibinfo{author}{McCormick, F.}, \bibinfo{author}{Gibeau, B.} \&
  \bibinfo{author}{Ghaemi, S.}
\newblock \bibinfo{title}{Reactive control of velocity fluctuations using an
  active deformable surface and real-time {PIV}}.
\newblock \emph{\bibinfo{journal}{Journal of Fluid Mechanics}}
  \textbf{\bibinfo{volume}{985}}, \bibinfo{pages}{A9} (\bibinfo{year}{2024}).

\bibitem[D{\'i}az et~al.(1983)]{diaz1983vortex}
\bibinfo{author}{D{\'i}az, F.}, \bibinfo{author}{Gavald{\`a}, J.},
  \bibinfo{author}{Kawall, J.~G.}, \bibinfo{author}{Keffer, J.~F.} \&
  \bibinfo{author}{Giralt, F.}
\newblock \bibinfo{title}{Vortex shedding from a spinning cylinder}.
\newblock \emph{\bibinfo{journal}{The Physics of Fluids}}
  \textbf{\bibinfo{volume}{26}}, \bibinfo{pages}{3454--3460}
  (\bibinfo{year}{1983}).

\bibitem[Aljure et~al.(2015)]{aljure2015influence}
\bibinfo{author}{Aljure, D.~E.}, \bibinfo{author}{Rodr{\'\i}guez, I.},
  \bibinfo{author}{Lehmkuhl, O.}, \bibinfo{author}{P{\'e}rez-Segarra, C.~D.} \&
  \bibinfo{author}{Oliva, A.}
\newblock \bibinfo{title}{Influence of rotation on the flow over a cylinder at
  {Re} = 5000}.
\newblock \emph{\bibinfo{journal}{International Journal of Heat and Fluid
  Flow}} \textbf{\bibinfo{volume}{55}}, \bibinfo{pages}{76--90}
  (\bibinfo{year}{2015}).

\bibitem[Acheson(1990)]{acheson1990elementary}
\bibinfo{author}{Acheson, D.~J.}
\newblock \emph{\bibinfo{title}{Elementary fluid dynamics}}
  (\bibinfo{publisher}{Oxford University Press}, \bibinfo{year}{1990}).

\bibitem[Banelli et~al.(2026)]{terpin2026flow}
\bibinfo{author}{Banelli, F.}, \bibinfo{author}{Terpin, A.},
  \bibinfo{author}{Bonomi, A.} \& \bibinfo{author}{D'Andrea, R.}
\newblock \bibinfo{title}{{Flow Gym}: A framework for the development,
  benchmarking, training, and deployment of flow-field quantification methods}.
\newblock \emph{\bibinfo{journal}{SoftwareX}} \textbf{\bibinfo{volume}{34}},
  \bibinfo{pages}{102641} (\bibinfo{year}{2026}).

\bibitem[Terpin et~al.(2026)]{terpin2026synthpix}
\bibinfo{author}{Terpin, A.}, \bibinfo{author}{Bonomi, A.},
  \bibinfo{author}{Banelli, F.} \& \bibinfo{author}{D'Andrea, R.}
\newblock \bibinfo{title}{{SynthPix}: A lightspeed {PIV} image generator}.
\newblock \emph{\bibinfo{journal}{SoftwareX}} \textbf{\bibinfo{volume}{34}},
  \bibinfo{pages}{102642} (\bibinfo{year}{2026}).

\bibitem[Badr et~al.(1990)]{badr1990unsteady}
\bibinfo{author}{Badr, H.~M.}, \bibinfo{author}{Coutanceau, M.},
  \bibinfo{author}{Dennis, S. C.~R.} \& \bibinfo{author}{M{\'e}nard, C.}
\newblock \bibinfo{title}{Unsteady flow past a rotating circular cylinder at
  {Reynolds} numbers {$10^3$} and {$10^4$}}.
\newblock \emph{\bibinfo{journal}{Journal of Fluid Mechanics}}
  \textbf{\bibinfo{volume}{220}}, \bibinfo{pages}{459--484}
  (\bibinfo{year}{1990}).

\bibitem[{\AA}str{\"o}m et~al.(2005)]{aastrom2005bicycle}
\bibinfo{author}{{\AA}str{\"o}m, K.~J.}, \bibinfo{author}{Klein, R.~E.} \&
  \bibinfo{author}{Lennartsson, A.}
\newblock \bibinfo{title}{Bicycle dynamics and control: Adapted bicycles for
  education and research}.
\newblock \emph{\bibinfo{journal}{IEEE Control Systems Magazine}}
  \textbf{\bibinfo{volume}{25}}, \bibinfo{pages}{26--47}
  (\bibinfo{year}{2005}).

\bibitem[Dallali et~al.(2009)]{dallali2009using}
\bibinfo{author}{Dallali, H.}, \bibinfo{author}{Brown, M.} \&
  \bibinfo{author}{Vanderborght, B.}
\newblock \bibinfo{title}{ in \textit{Using the torso to compensate for
  non-minimum phase behaviour in {ZMP} bipedal walking}}
  \emph{\bibinfo{booktitle}{Advances in Robotics Research: Theory,
  Implementation, Application}} \bibinfo{pages}{191--202}
  (\bibinfo{publisher}{Springer Berlin Heidelberg}, \bibinfo{year}{2009}).

\bibitem[Wee et~al.(2013)]{wee2013design}
\bibinfo{author}{Wee, T.-C.}, \bibinfo{author}{Astolfi, A.} \&
  \bibinfo{author}{Xie, M.}
\newblock \bibinfo{title}{The design and control of a bipedal robot with
  sensory feedback}.
\newblock \emph{\bibinfo{journal}{International Journal of Advanced Robotic
  Systems}} \textbf{\bibinfo{volume}{10}}, \bibinfo{pages}{277}
  (\bibinfo{year}{2013}).

\bibitem[Hauser et~al.(1992)]{hauser1992nonlinear}
\bibinfo{author}{Hauser, J.}, \bibinfo{author}{Sastry, S.} \&
  \bibinfo{author}{Meyer, G.}
\newblock \bibinfo{title}{Nonlinear control design for slightly non-minimum
  phase systems: Application to {V/STOL} aircraft}.
\newblock \emph{\bibinfo{journal}{Automatica}} \textbf{\bibinfo{volume}{28}},
  \bibinfo{pages}{665--679} (\bibinfo{year}{1992}).

\bibitem[Tomlin et~al.(1995)]{tomlin1995output}
\bibinfo{author}{Tomlin, C.}, \bibinfo{author}{Lygeros, J.},
  \bibinfo{author}{Benvenuti, L.} \& \bibinfo{author}{Sastry, S.}
  \emph{\bibinfo{title}{Output tracking for a non-minimum phase dynamic {CTOL}
  aircraft model}}.
\newblock \emph{\bibinfo{booktitle}{Proceedings of the 34th IEEE Conference on
  Decision and Control}}, Vol.~\bibinfo{volume}{2}, \bibinfo{pages}{1867--1872}
  (\bibinfo{publisher}{IEEE}, \bibinfo{year}{1995}).

\bibitem[Agyei et~al.(2026)]{agyei2026deep}
\bibinfo{author}{Agyei, K.}, \bibinfo{author}{Sarhadi, P.} \&
  \bibinfo{author}{Polani, D.}
\newblock \bibinfo{title}{Deep reinforcement learning in applied control:
  Challenges, analysis, and insights}.
\newblock \emph{\bibinfo{journal}{Engineering Applications of Artificial
  Intelligence}} \textbf{\bibinfo{volume}{181}}, \bibinfo{pages}{115524}
  (\bibinfo{year}{2026}).

\bibitem[Charfeddine et~al.(2025)]{charfeddine2025adaptive}
\bibinfo{author}{Charfeddine, M.}, \bibinfo{author}{Jouili, K.} \&
  \bibinfo{author}{Ben~Moussa, M.}
\newblock \bibinfo{title}{Adaptive control of nonlinear non-minimum phase
  systems using actor--critic reinforcement learning}.
\newblock \emph{\bibinfo{journal}{Symmetry}} \textbf{\bibinfo{volume}{17}},
  \bibinfo{pages}{1083} (\bibinfo{year}{2025}).

\bibitem[van Zundert and Oomen(2018)]{van2018inversion}
\bibinfo{author}{van Zundert, J.} \& \bibinfo{author}{Oomen, T.}
\newblock \bibinfo{title}{On inversion-based approaches for feedforward and
  {ILC}}.
\newblock \emph{\bibinfo{journal}{Mechatronics}} \textbf{\bibinfo{volume}{50}},
  \bibinfo{pages}{282--291} (\bibinfo{year}{2018}).

\bibitem[Zhou et~al.(2018)]{zhou2018inversion}
\bibinfo{author}{Zhou, S.}, \bibinfo{author}{Helwa, M.~K.} \&
  \bibinfo{author}{Schoellig, A.~P.}
\newblock \bibinfo{title}{An inversion-based learning approach for improving
  impromptu trajectory tracking of robots with non-minimum phase dynamics}.
\newblock \emph{\bibinfo{journal}{IEEE Robotics and Automation Letters}}
  \textbf{\bibinfo{volume}{3}}, \bibinfo{pages}{1663--1670}
  (\bibinfo{year}{2018}).

\bibitem[Doyle(1978)]{doyle1978guaranteed}
\bibinfo{author}{Doyle, J.~C.}
\newblock \bibinfo{title}{Guaranteed margins for {LQG} regulators}.
\newblock \emph{\bibinfo{journal}{IEEE Transactions on Automatic Control}}
  \textbf{\bibinfo{volume}{23}}, \bibinfo{pages}{756--757}
  (\bibinfo{year}{1978}).

\bibitem[Leong and Doyle(2016)]{leong2016understanding}
\bibinfo{author}{Leong, Y.~P.} \& \bibinfo{author}{Doyle, J.~C.}
  \emph{\bibinfo{title}{Understanding robust control theory via stick
  balancing}}.
\newblock \emph{\bibinfo{booktitle}{2016 IEEE 55th Conference on Decision and
  Control (CDC)}}, \bibinfo{pages}{1508--1514} (\bibinfo{organization}{IEEE},
  \bibinfo{year}{2016}).

\bibitem[Xu et~al.(2021)]{xu2021learned}
\bibinfo{author}{Xu, J.}, \bibinfo{author}{Lee, B.}, \bibinfo{author}{Matni,
  N.} \& \bibinfo{author}{Jayaraman, D.} \emph{\bibinfo{title}{How are learned
  perception-based controllers impacted by the limits of robust control?}}
\newblock \emph{\bibinfo{booktitle}{Proceedings of the 3rd Conference on
  Learning for Dynamics and Control}}, Vol. \bibinfo{volume}{144} of
  \emph{\bibinfo{series}{Proceedings of Machine Learning Research}},
  \bibinfo{pages}{954--966} (\bibinfo{publisher}{PMLR}, \bibinfo{year}{2021}).

\bibitem[Lee et~al.(2024)]{lee2024active}
\bibinfo{author}{Lee, B.~D.}, \bibinfo{author}{Ziemann, I.},
  \bibinfo{author}{Pappas, G.~J.} \& \bibinfo{author}{Matni, N.}
  \emph{\bibinfo{title}{Active learning for control-oriented identification of
  nonlinear systems}}.
\newblock \emph{\bibinfo{booktitle}{2024 IEEE 63rd Conference on Decision and
  Control (CDC)}}, \bibinfo{pages}{3011--3018} (\bibinfo{organization}{IEEE},
  \bibinfo{year}{2024}).

\bibitem[Baumann et~al.(2018)]{baumann2018deep}
\bibinfo{author}{Baumann, D.}, \bibinfo{author}{Zhu, J.-J.},
  \bibinfo{author}{Martius, G.} \& \bibinfo{author}{Trimpe, S.}
  \emph{\bibinfo{title}{Deep reinforcement learning for event-triggered
  control}}.
\newblock \emph{\bibinfo{booktitle}{2018 IEEE Conference on Decision and
  Control (CDC)}}, \bibinfo{pages}{943--950} (\bibinfo{organization}{IEEE},
  \bibinfo{year}{2018}).

\bibitem[Nam et~al.(2021)]{nam2021reinforcement}
\bibinfo{author}{Nam, H.~A.}, \bibinfo{author}{Fleming, S.} \&
  \bibinfo{author}{Brunskill, E.} \emph{\bibinfo{title}{Reinforcement learning
  with state observation costs in action-contingent noiselessly observable
  {Markov} decision processes}}.
\newblock \emph{\bibinfo{booktitle}{Advances in Neural Information Processing
  Systems}}, Vol.~\bibinfo{volume}{34}, \bibinfo{pages}{15650--15666}
  (\bibinfo{year}{2021}).

\bibitem[Holt et~al.(2023)]{holt2023active}
\bibinfo{author}{Holt, S.}, \bibinfo{author}{H{\"u}y{\"u}k, A.} \&
  \bibinfo{author}{van~der Schaar, M.} \emph{\bibinfo{title}{Active observing
  in continuous-time control}}.
\newblock \emph{\bibinfo{booktitle}{Advances in Neural Information Processing
  Systems}}, Vol.~\bibinfo{volume}{36}, \bibinfo{pages}{46054--46092}
  (\bibinfo{year}{2023}).

\bibitem[Brockett(1997)]{brockett1997minimum}
\bibinfo{author}{Brockett, R.~W.} \emph{\bibinfo{title}{Minimum attention
  control}}.
\newblock \emph{\bibinfo{booktitle}{Proceedings of the 36th IEEE Conference on
  Decision and Control}}, Vol.~\bibinfo{volume}{3}, \bibinfo{pages}{2628--2632}
  (\bibinfo{organization}{IEEE}, \bibinfo{year}{1997}).

\bibitem[Sailer et~al.(2005)]{sailer2005eye}
\bibinfo{author}{Sailer, U.}, \bibinfo{author}{Flanagan, J.~R.} \&
  \bibinfo{author}{Johansson, R.~S.}
\newblock \bibinfo{title}{Eye--hand coordination during learning of a novel
  visuomotor task}.
\newblock \emph{\bibinfo{journal}{The Journal of Neuroscience}}
  \textbf{\bibinfo{volume}{25}}, \bibinfo{pages}{8833--8842}
  (\bibinfo{year}{2005}).

\bibitem[Wang et~al.(2021)]{wang2021experimental}
\bibinfo{author}{Wang, X.}, \bibinfo{author}{Chen, J.}, \bibinfo{author}{Zhou,
  B.}, \bibinfo{author}{Li, Y.} \& \bibinfo{author}{Xiang, Q.}
\newblock \bibinfo{title}{Experimental investigation of flow past a confined
  bluff body: Effects of body shape, blockage ratio and {Reynolds} number}.
\newblock \emph{\bibinfo{journal}{Ocean Engineering}}
  \textbf{\bibinfo{volume}{220}}, \bibinfo{pages}{108412}
  (\bibinfo{year}{2021}).

\bibitem[Lu et~al.(2025)]{lu2025spectral}
\bibinfo{author}{Lu, W.}, \bibinfo{author}{Chan, L.} \& \bibinfo{author}{Ooi,
  A.}
\newblock \bibinfo{title}{Spectral analysis of confined cylinder wakes}.
\newblock \emph{\bibinfo{journal}{Fluids}} \textbf{\bibinfo{volume}{10}},
  \bibinfo{pages}{84} (\bibinfo{year}{2025}).

\bibitem[Rabault and Kuhnle(2019)]{rabault2019accelerating}
\bibinfo{author}{Rabault, J.} \& \bibinfo{author}{Kuhnle, A.}
\newblock \bibinfo{title}{Accelerating deep reinforcement learning strategies
  of flow control through a multi-environment approach}.
\newblock \emph{\bibinfo{journal}{Physics of Fluids}}
  \textbf{\bibinfo{volume}{31}}, \bibinfo{pages}{094105}
  (\bibinfo{year}{2019}).

\bibitem[Raffel et~al.(2018)]{raffel2018particle}
\bibinfo{author}{Raffel, M.}, \bibinfo{author}{Willert, C.~E.},
  \bibinfo{author}{Scarano, F.}, \bibinfo{author}{K{\"a}hler, C.~J.},
  \bibinfo{author}{Wereley, S.~T.} \& \bibinfo{author}{Kompenhans, J.}
\newblock \emph{\bibinfo{title}{Particle image velocimetry: a practical guide}}
  \bibinfo{edition}{3} edn (\bibinfo{publisher}{Springer International
  Publishing}, \bibinfo{address}{Cham}, \bibinfo{year}{2018}).

\bibitem[Kroeger et~al.(2016)]{kroeger2016fast}
\bibinfo{author}{Kroeger, T.}, \bibinfo{author}{Timofte, R.},
  \bibinfo{author}{Dai, D.} \& \bibinfo{author}{Van~Gool, L.}
  \emph{\bibinfo{title}{Fast optical flow using dense inverse search}}.
\newblock \emph{\bibinfo{booktitle}{Computer Vision---{ECCV} 2016}}, Vol.
  \bibinfo{volume}{9908} of \emph{\bibinfo{series}{Lecture Notes in Computer
  Science}}, \bibinfo{pages}{471--488} (\bibinfo{publisher}{Springer, Cham},
  \bibinfo{year}{2016}).

\bibitem[Olson(2011)]{olson2011apriltag}
\bibinfo{author}{Olson, E.} \emph{\bibinfo{title}{{AprilTag}: A robust and
  flexible visual fiducial system}}.
\newblock \emph{\bibinfo{booktitle}{Proceedings of the {IEEE} International
  Conference on Robotics and Automation ({ICRA})}}, \bibinfo{pages}{3400--3407}
  (\bibinfo{publisher}{IEEE}, \bibinfo{year}{2011}).

\bibitem[Norberg(1987)]{norberg1987effect}
\bibinfo{author}{Norberg, C.}
\newblock \bibinfo{title}{Effects of {Reynolds} number and a low-intensity
  freestream turbulence on the flow around a circular cylinder}.
\newblock \bibinfo{type}{Publikation} \bibinfo{number}{87/2},
  \bibinfo{institution}{Chalmers University of Technology},
  \bibinfo{address}{Gothenburg, Sweden} (\bibinfo{year}{1987}).

\bibitem[Mondal and Alam(2023)]{mondal2023blockage}
\bibinfo{author}{Mondal, R.} \& \bibinfo{author}{Alam, M.~M.}
\newblock \bibinfo{title}{Blockage effect on wakes of various bluff bodies: A
  review of confined flow}.
\newblock \emph{\bibinfo{journal}{Ocean Engineering}}
  \textbf{\bibinfo{volume}{286}}, \bibinfo{pages}{115592}
  (\bibinfo{year}{2023}).

\bibitem[{Microchip Technology Inc.}(2024)]{microchip2024sine}
\bibinfo{author}{{Microchip Technology Inc.}}
\newblock \bibinfo{title}{Sine wave generator using numerically controlled
  oscillator module}.
\newblock \bibinfo{type}{Application Note} \bibinfo{number}{AN1523},
  \bibinfo{institution}{Microchip Technology Inc.} (\bibinfo{year}{2024}).
\newblock \bibinfo{note}{Document DS00001523B, Revision B}.

\bibitem[Schulman
  et~al.(2017)]{schulman2017proximalpolicyoptimizationalgorithms}
\bibinfo{author}{Schulman, J.}, \bibinfo{author}{Wolski, F.},
  \bibinfo{author}{Dhariwal, P.}, \bibinfo{author}{Radford, A.} \&
  \bibinfo{author}{Klimov, O.}
\newblock \bibinfo{title}{Proximal policy optimization algorithms}
  (\bibinfo{year}{2017}).
\newblock \bibinfo{note}{ArXiv:1707.06347}.

\bibitem[Haarnoja et~al.(2018)]{haarnoja2018softactorcriticoffpolicymaximum}
\bibinfo{author}{Haarnoja, T.}, \bibinfo{author}{Zhou, A.},
  \bibinfo{author}{Abbeel, P.} \& \bibinfo{author}{Levine, S.}
  \emph{\bibinfo{title}{Soft actor-critic: Off-policy maximum entropy deep
  reinforcement learning with a stochastic actor}}.
\newblock \emph{\bibinfo{booktitle}{Proceedings of the 35th International
  Conference on Machine Learning}}, Vol.~\bibinfo{volume}{80} of
  \emph{\bibinfo{series}{Proceedings of Machine Learning Research}},
  \bibinfo{pages}{1861--1870} (\bibinfo{publisher}{PMLR},
  \bibinfo{year}{2018}).

\bibitem[Raffin et~al.(2021)]{raffin2021stable}
\bibinfo{author}{Raffin, A.}, \bibinfo{author}{Hill, A.},
  \bibinfo{author}{Gleave, A.}, \bibinfo{author}{Kanervisto, A.},
  \bibinfo{author}{Ernestus, M.} \& \bibinfo{author}{Dormann, N.}
\newblock \bibinfo{title}{{Stable-Baselines3}: Reliable reinforcement learning
  implementations}.
\newblock \emph{\bibinfo{journal}{Journal of Machine Learning Research}}
  \textbf{\bibinfo{volume}{22}}, \bibinfo{pages}{1--8} (\bibinfo{year}{2021}).

\bibitem[Welch(1967)]{welch1967power}
\bibinfo{author}{Welch, P.~D.}
\newblock \bibinfo{title}{The use of fast {Fourier} transform for the
  estimation of power spectra: A method based on time averaging over short,
  modified periodograms}.
\newblock \emph{\bibinfo{journal}{IEEE Transactions on Audio and
  Electroacoustics}} \textbf{\bibinfo{volume}{15}}, \bibinfo{pages}{70--73}
  (\bibinfo{year}{1967}).

\bibitem[Lin(1991)]{lin1991divergence}
\bibinfo{author}{Lin, J.}
\newblock \bibinfo{title}{Divergence measures based on the {Shannon} entropy}.
\newblock \emph{\bibinfo{journal}{IEEE Transactions on Information Theory}}
  \textbf{\bibinfo{volume}{37}}, \bibinfo{pages}{145--151}
  (\bibinfo{year}{1991}).

\bibitem[Whitehead and Ballard(1991)]{whitehead1991learning}
\bibinfo{author}{Whitehead, S.~D.} \& \bibinfo{author}{Ballard, D.~H.}
\newblock \bibinfo{title}{Learning to perceive and act by trial and error}.
\newblock \emph{\bibinfo{journal}{Machine Learning}}
  \textbf{\bibinfo{volume}{7}}, \bibinfo{pages}{45--83} (\bibinfo{year}{1991}).

\end{thebibliography}

\clearpage
\begin{appendices}
\section{Methods}
\label{sec:methods}
\subsection[Experimental apparatus and measurement pipeline]{Experimental apparatus and measurement pipeline}
\label{sec:system:water-channel}
We collect data on an Ubuntu 22.04 machine with an AMD Ryzen Threadripper PRO 5995WX and \PFFProtocolComputeGPUCount{} Nvidia RTX 4090 GPUs. A dedicated GPU runs the flow estimation pipeline; separate dedicated GPUs run policy inference and training.

\cref{fig:hardware_overview} illustrates the tabletop \gls*{cwc}, with close-ups and labels for all components.
We regulate the mean incoming flow speed at the beginning of the test section to $\PFFProtocolIncomingFlowSpeedMPerS\,\mathrm{m}/\mathrm{s}$.
We list all off-the-shelf parts below and 3D-print the remaining components. Their designs and fabrication files are available at \url{https://activefluidcontrol.com}.
The setup fits inside a $\PFFProtocolAquariumLengthM \times \PFFProtocolAquariumWidthM \times \PFFProtocolAquariumHeightM\,\mathrm{m}^3$ Long Aquarium ($\PFFProtocolAquariumNominalVolumeL\,\mathrm{L}$, Olibetta), with a test section of $\PFFProtocolTestSectionLengthM \times \PFFProtocolTestSectionWidthM \times \PFFProtocolTestSectionHeightM\,\mathrm{m}^3$.

The channel consists of an upper and a lower branch. The test section (\textbf{1}) in the upper branch houses the cylinder (\textbf{2}). \PFFProtocolPropellerCount{} APISQUEEN U5 brushless propellers (\textbf{3}) drive a left-to-right flow; guide vanes (\textbf{4}) redirect the flow into the lower branch and then back to the upper branch. A honeycomb structure (\textbf{5}) straightens the flow and suppresses large-scale vortices. Flow restrictions (\textbf{6}) accelerate the stream and stretch remaining small-scale vortices. The \gls*{piv} setup (\textbf{7}) captures images at $\PFFProtocolPIVCameraRateHz\,\mathrm{Hz}$; the flow estimation pipeline processes consecutive image pairs to produce flow field estimates in real time \citep{terpin2026flow}.

We attach the cylinder (\textbf{2}), the setup's controlled object, to a Maxon EC45 flat motor with a \PFFProtocolEncoderPPR{} PPR incremental encoder. The cylinder radius is \(R = \PFFProtocolCylinderRadiusM\,\mathrm{m}\). We cap the rotation rate at \(\bar{A} = \PFFProtocolMaximumRotationRateRadPerS\,\mathrm{rad/s}\), which corresponds to imposing a surface flow speed of about \(\PFFProtocolMaximumSurfaceSpeedMPerSRoundedTwo\,\mathrm{m/s}\), more than twice the average incoming flow speed, so that the actuation can locally reverse the flow. Using the bounds \(\mathrm{St} \leq \PFFProtocolStrouhalUpperBound\) for the Strouhal number \citep{wang2021experimental,lu2025spectral} and \(U_{\infty} \leq \PFFProtocolInflowSpeedUpperBoundMPerS\,\mathrm{m/s}\), together with \(f_{\mathcal{C}} = \mathrm{St} U_{\infty}/D\), we obtain the crude characteristic-frequency bound \(f_{\mathcal{C}} \leq \PFFProtocolControlCharacteristicFrequencyHzRoundedOne\,\mathrm{Hz}\). We therefore adopt a control-input update rate of \(\PFFProtocolSampleRateHz\,\mathrm{Hz}\), providing more than ten updates per period at this bound. This update ratio is consistent with the action timescale in a prior \gls*{rl} study of cylinder-wake control \citep{rabault2019accelerating}.

We mount the motor on a shaft at $\vec{x}_0$, parallel to the flow surface and transverse to the incoming flow. A TD50 sensor by ME-Meßsysteme GmbH (\textbf{2}) measures torque about this shaft, with full scale $\PFFProtocolTorqueSensorFullScaleMNm\,\mathrm{mNm}$ and accuracy $\PFFProtocolTorqueSensorAccuracyPercentFullScale\%$ of full scale.
The measured signal \(\hat{\tau}_{\text{drag}}^m(t)\) is a noisy estimate of the torque \(\tau_{\text{drag}}(t)\) and an indirect measurement of the drag force \(F_D(t)\) in the direction of the flow \(\vec{n}_{\infty}\). We assume that the unchanged mounting geometry fixes an effective lever arm \(\ell\), such that \(\tau_{\text{drag}}(t)=\ell F_D(t)\) in every experimental condition. The constant \(\ell\) cancels from all relative changes. Accordingly, throughout the paper, a ``drag measurement'' denotes this torque-proportional proxy; we use ``torque'' only when discussing the sensor signal or its units.
We sample $\hat{\tau}_{\text{drag}}^m(t)$ at $\PFFProtocolTorqueSampleRateHzDisplay\,\mathrm{Hz}$ and average the signal over the preceding $\PFFProtocolTorqueFilterWindowS\,\mathrm{s}$ to reduce sensor noise and vortex-shedding fluctuations. The reward and reported drag proxy therefore use this filtered signal, and we evaluate episode performance over $T=\PFFProtocolEpisodeDurationS\,\mathrm{s}$. We characterize actuation-induced measurement effects via a separate \hyperref[sec:water-channel:torque]{experiment without incoming flow}.
We define the common normalization reference as the mean of the \PFFProtocolBaselineGridRepetitions{} trial means for the no-control condition with incoming flow, $A=\PFFProtocolBaselineNoControlAmplitudeRadPerS\,\mathrm{rad/s}$ and $f=\PFFProtocolBaselineGridFrequencyMinHz\,\mathrm{Hz}$; see \nameref{sec:experiments:open-loop}. Each trial mean uses the full \PFFProtocolEpisodeDurationS-second filtered-drag history. The resulting reference is $\PFFExperimentNoControlReportingMNmRoundedOne\,\mathrm{mNm}$.

To obtain flow observations, we use \gls*{piv}, a standard method for estimating velocity fields in bluff-body wakes \citep{raffel2018particle}. Our setup uses the Optolution \gls*{piv} system (\textbf{7}).
In \gls*{piv}, laser pulses illuminate neutrally buoyant particles in the fluid through the transparent acrylic section in the middle of the flow restrictions at specific intervals, while the camera captures images. We use \(\PFFProtocolPIVTracerDiameterUm\,\mathrm{\mu m}\)-diameter hollow glass spheres (LaVision) as tracer particles.
\Cref{fig:piv-timing} depicts the timing sequence of camera exposure (blue line, top) and laser pulses (red line, below). The bottom part of the figure shows two \gls*{piv} images that the camera captures at $t_0$ and $t_0+\Delta t$. In our setup, each image contains approximately $\PFFProtocolPIVCroppedImageMegapixels$ megapixel. The flow estimation pipeline uses these pairs to track particle displacement and produce flow field estimates.
\Cref{fig:hardware_overview} shows examples of the resulting flow field estimates.
\PFFFigure[.4\linewidth]{8}

While the Optolution \gls*{piv} system yields high-quality, high-resolution image pairs, it does not provide real-time processing and, more importantly, off-the-shelf open-source options fail to deliver vectors that are accurate enough for flow estimation while processing high-resolution images at high frequency. We therefore adapted the Dense Inverse Search algorithm from \citet{kroeger2016fast} to run entirely on the GPU. The extra speedup enabled us to tune the algorithm so as to obtain measurements that are accurate enough for our purposes. For tuning, we also developed a synthetic image generator that mimics our optical setup and seeding conditions. We describe these developments in detail, and their relevance for the fluid-control community, in \citep{terpin2026flow,terpin2026synthpix}.
Concretely, we mark the region of interest in the wake of the cylinder using AprilTags \citep{olson2011apriltag}. We leverage the detected tags for automatic camera calibration. The flow estimation pipeline crops full-resolution frames to a $\sim\PFFProtocolPIVCroppedImageMegapixels$-megapixel window covering this region, resizes each crop to $\PFFProtocolPIVProcessingGridRows\times\PFFProtocolPIVProcessingGridColumns$, and processes it to yield a $\PFFProtocolPIVProcessingGridRows\times\PFFProtocolPIVProcessingGridColumns$ flow estimate. After estimating the flow field, we can linearly downsample it to as few as $\PFFProtocolFlowObservationGridRows\times\PFFProtocolFlowObservationGridColumns$ grid points without appreciable loss of salient information (instead, inference on lower-resolution images substantially degrades the estimate).
This is practically interesting because $\PFFProtocolFlowObservationGridRows\times\PFFProtocolFlowObservationGridColumns$ flow estimates are higher-dimensional than the feedback that the fluids-control literature typically uses, yet remain very amenable to state-of-the-art \gls*{rl} methods.

\subsection{Additional sources of uncertainty in the experiments}
\label{sec:uncertainties}
In our experiments, we recognize that several secondary sources of uncertainty can influence the data. For instance, the uncertainties due to the torque sensor measurements are below $\PFFProtocolTorqueSensorUncertaintyUpperMNmRoundedOne\,\mathrm{mNm}$ and the ones related to non-idealities in the fabrication and assembly are below $\PFFProtocolFabricationAssemblyUncertaintyUpperMNm\,\mathrm{mNm}$ for the cylinder. Other sources of uncertainty include natural variations in room temperature, small fluctuations in the water level, and minor mechanical non-idealities, such as slightly different tracking performance on the desired speed of the propellers. These effects may introduce small discrepancies in the results, but their overall effect is limited. We use $\PFFProtocolMaterialEffectThresholdMNm\,\mathrm{mNm}$ ($\PFFPaperMaterialEffectThresholdPercent\%$ of the no-control mean) as the material-effect threshold; this threshold exceeds the combined sensor, fabrication, and assembly uncertainty. Finally, since we collect data over multiple weeks, the no-control baseline shows minor variations across experiments. This bias, for instance due to temperature changes, is minor, and we systematically remove it by considering changes in torque measurements rather than their absolute values.

At $\mathrm{Re}=U_\infty D/\nu\approx\PFFProtocolReynoldsNumberDisplay$, the wake (\cref{fig:cover}) is in the subcritical vortex-shedding regime \citep{norberg1987effect}. The blockage ratio $D/W\approx\PFFProtocolBlockageRatioRoundedTwo$, where $W=\PFFProtocolTestSectionWidthM\,\mathrm{m}$ is the test-section width, produces strong confinement that materially affects the wake \citep{mondal2023blockage}. This characteristic of our system is both a feature and a limitation. It is a feature because the strong confinement 
complicates the flow dynamics, introduces additional effects and increases the difficulty of simulation as well as the potential of feedback control. It is a limitation because the results may not generalize to unconfined flows.

\subsection{Additional experiment to verify that the cylinder dynamics are inconsequential for the torque sensor measurements}
\label{sec:water-channel:torque}
To verify that actuation itself does not contaminate our torque-sensor measurements, we performed a dedicated characterization. Because the motor and sensor share the same shaft, the motor dynamics together with misalignment (due to, e.g., fabrication inaccuracies) of the center of mass of the cylinder with respect to the rotational axis could, in principle, affect the measured torque. We therefore designed a simple experiment to bound any actuation-induced artifacts under representative operating conditions.
In a quiescent, no-flow environment, we command the cylinder
to track the constant and sinusoidal open-loop profiles over the same grid as in \nameref{sec:experiments:open-loop}. For each grid point, we collect $\PFFProtocolEpisodeDurationS\,\mathrm{s}$ of data and repeat the experiment $\PFFProtocolBaselineMotorCharacterizationRepetitions$ times.
Overall, the average measured torque ranges approximately from $\PFFPaperMotorImpactMinimumMNm$ to $\PFFPaperMotorImpactMaximumMNm\,\mathrm{mNm}$. This range spans $\PFFPaperMotorImpactSpanPercent\%$ of the $\PFFExperimentNoControlReportingMNmRoundedOne\,\mathrm{mNm}$ no-control reporting reference and is small relative to the drag changes in \nameref{sec:results}.

\subsection[Open-loop sinusoidal reference selection]{Open-loop sinusoidal reference selection}
\label{sec:experiments:open-loop}
We describe our procedure to verify that simple sinusoidal open-loop inputs can achieve high performance and to identify representative high-performing sinusoids for both drag maximization and drag minimization. Note that, for the scope of our work and claims, it is not important to find the globally best open-loop policy, or even the best sinusoidal one; it suffices to obtain a high-performing sinusoidal reference.
We systematically probe the system with constant inputs $\omega(t)=A$ and, for $f>0$, sinusoidal inputs $\omega(t)=A\sin(2\pi ft)$.
At $f=0$, we test the constant rotation rates
$
A\in\{\PFFProtocolBaselineConstantAmplitudesRadPerS\}\,\mathrm{rad}/\mathrm{s}.
$
At positive frequencies $f\in[\PFFProtocolBaselinePositiveFrequencyMinHz,\PFFProtocolBaselineGridFrequencyMaxHz]\,\mathrm{Hz}$, we sample frequencies in increments of $\PFFProtocolBaselineGridFrequencyStepHz\,\mathrm{Hz}$ and test the amplitudes
$A\in\{\PFFProtocolBaselineGridAmplitudesRadPerS\}\,\mathrm{rad}/\mathrm{s}$.
The no-control baseline is the constant condition $A=\PFFProtocolBaselineNoControlAmplitudeRadPerS$. The positive-frequency range and our $\PFFProtocolSampleRateHz\,\mathrm{Hz}$ actuation rate ensure at least \PFFProtocolBaselineMinimumControlUpdatesPerPeriod{} control updates per period \citep{microchip2024sine}.
This gives $\PFFProtocolBaselineGridSettings$ tested settings: nine constant-rate conditions and $4\times\PFFProtocolBaselinePositiveFrequencyConditions$ sinusoidal conditions. We obtain the commanded amplitude setpoints by unit conversion. For each setting, we collect \PFFProtocolBaselineGridRepetitions{} repetitions and compute the mean performance, observing fairly consistent outcomes across repetitions.
This first search suggests the candidate $(A = \PFFProtocolBaselineSelectedAmplitudeRadPerS\,\mathrm{rad}/\mathrm{s}, f = \PFFProtocolBaselineGridSelectedMaxFrequencyHz\,\mathrm{Hz})$ for drag maximization; the numerical minimum for drag minimization occurs at $(A = \PFFProtocolBaselineSelectedAmplitudeRadPerS\,\mathrm{rad}/\mathrm{s}, f = \PFFPaperBaselineGridNumericalMinimumFrequencyHz\,\mathrm{Hz})$, with performance plateauing for $f > \PFFProtocolBaselineGridPlateauOnsetHz\,\mathrm{Hz}$. The numerical minimizing condition lies adjacent to the upper boundary of the initial scan. We select the tested boundary at $f=\PFFProtocolBaselineGridSelectedMinFrequencyHz\,\mathrm{Hz}$ as the high-performing drag-minimizing sinusoidal reference. At higher frequencies, performance eventually degrades because of tracking and sampling limitations. We also find the highest-performing amplitude at the upper boundary, $A = \PFFProtocolBaselineSelectedAmplitudeRadPerS\,\mathrm{rad}/\mathrm{s}$, and impose the same physically motivated limit on the \gls*{rl} agent; see \nameref{sec:system:water-channel}. Moreover, increasing the amplitude further leads to action trajectories that excite strong surface waves in the water channel and amplify boundary effects, so we do not consider larger amplitudes.
Overall, we use the grid means at \((A = \PFFProtocolBaselineSelectedAmplitudeRadPerS\,\mathrm{rad/s},\, f = \PFFProtocolBaselineGridSelectedMaxFrequencyHz\,\mathrm{Hz})\) for drag maximization and \((A = \PFFProtocolBaselineSelectedAmplitudeRadPerS\,\mathrm{rad/s},\, f = \PFFProtocolBaselineGridSelectedMinFrequencyHz\,\mathrm{Hz})\) for drag minimization as the high-performing sinusoidal references. We compute each reference from \PFFProtocolBaselineGridRepetitions{} physical repetitions. Relative to the common no-control reference, the drag-maximizing reference produces an increase of \(\PFFPaperBaselineMaxReferencePercentPM\%\), and the drag-minimizing reference produces a decrease of \(\PFFPaperBaselineMinReferencePercentPM\%\); the plus--minus terms are sample standard deviations across repetitions. Applying the small-sample Student \(t\) correction \(\bar{x}\mathbin{\pm}t_{\PFFProtocolStudentTQuantileProbability,\PFFProtocolBaselineStudentTDegreesFreedom}s/\sqrt{\PFFProtocolBaselineGridRepetitions{}}\), with \(t_{\PFFProtocolStudentTQuantileProbability,\PFFProtocolBaselineStudentTDegreesFreedom}=\PFFPaperBaselineStudentTCriticalValue\), gives two-sided \PFFProtocolConfidenceLevelPercent\% confidence intervals for mean performance of \([\PFFPaperBaselineMaxReferenceConfidenceLowerPercent,\PFFPaperBaselineMaxReferenceConfidenceUpperPercent]\%\) and \([\PFFPaperBaselineMinReferenceConfidenceLowerPercent,\PFFPaperBaselineMinReferenceConfidenceUpperPercent]\%\), respectively. With only \PFFProtocolBaselineGridRepetitions{} repetitions, the normality assumption cannot be assessed reliably. These intervals are descriptive, pointwise uncertainty summaries conditional on the selected settings; they do not adjust for selection over the grid or propagate uncertainty in the common no-control normalization.

\subsection[Training scores and threshold attainment]{Training scores and threshold attainment}
\label{appendix:headline-training-metrics}

We orient both tasks so that a larger score denotes better performance. For task--observation condition $c$, training repetition $j$, and episode $e$, let $\overline d_{cje}$ be the episode-average filtered drag and $d^0_{cje}$ its first value, both expressed in $\mathrm{mNm}$. With $s_c=+1$ for drag maximization and $s_c=-1$ for drag minimization, the episode score is
$x_{cje}=s_c\left(\overline d_{cje}-d^0_{cje}\right)$.
Thus, positive scores denote a drag increase in the maximization task and a drag reduction in the minimization task. The average uses the full recorded episode.

Let $b_c$ be the magnitude of the mean drag change produced by the corresponding selected high-performing sinusoid; see \nameref{sec:experiments:open-loop}. With episode duration $T_{\mathrm{ep}}=\PFFProtocolEpisodeDurationS\,\mathrm{s}$, the first-attainment time at the fraction $\alpha$ of the high-performing sinusoidal baseline, in active-interaction minutes, is
$L_{cj}(\alpha)=\frac{T_{\mathrm{ep}}}{60\,\mathrm{s}}\min\left\{e+1:x_{cje}\geq \alpha b_c\right\}.$
We define $L_{cj}$ only when the set in the expression is nonempty and report it in active-interaction minutes, excluding the stabilization intervals. For the first-attainment summaries, we use \PFFProtocolReferenceAttainmentPercent\% of the mean drag-change magnitude produced by the selected high-performing sinusoid for the corresponding task: a \PFFPaperHeadlineMaxAttainmentDragChangePercent\% drag increase for maximization and a \PFFPaperHeadlineMinAttainmentDragChangePercent\% drag reduction for minimization, both relative to no control. When a repetition does not reach this level, we report non-attainment within the $\PFFProtocolTrainingBudgetMinutes$-minute budget. This is the first observed qualifying episode, not a unique instant at which learning occurs.

Final performance is the mean score over the last $\PFFProtocolEndpointEpisodeCount{}$ episodes of each repetition:
$z_{cj}=\frac{1}{\PFFProtocolEndpointEpisodeCount}\sum_{e=\PFFProtocolEndpointFirstIndex}^{\PFFProtocolEndpointLastIndex}x_{cje}.$
For the incomplete records listed in \cref{appendix:experimental-accounting}, we divide by the number of available measurements or final-window episodes rather than the nominal counts; we do not impute missing values.

To express performance relative to the apparatus scale, we use the common no-control reference defined in \nameref{sec:system:water-channel}. It is the mean of the $\PFFProtocolBaselineGridRepetitions{}$ full-history trial means and equals $d_{\mathrm{nc}}\approx\PFFExperimentNoControlReportingMNmRoundedOne\,\mathrm{mNm}$. Expressing $z_{cj}$ and $d_{\mathrm{nc}}$ in the same units, the normalized endpoint is
$
q_{cj}=100\frac{z_{cj}}{d_{\mathrm{nc}}}.
$
An endpoint $q_{cj}$ is a percentage of the no-control reference; a difference between two such endpoints is measured in percentage points. The sinusoidal benchmarks $b_c$ and the no-control reference $d_{\mathrm{nc}}$ are fixed empirical quantities in these calculations, so the reported intervals do not propagate uncertainty in their estimation. We calculate attainment and final performance within each repetition before aggregating. Unless we explicitly give a confidence interval, values reported with a plus--minus sign are arithmetic means and sample standard deviations across independent training repetitions.
Training outcomes and threshold sensitivity are reported in \cref{tab:headline-training-summary,tab:headline-training-threshold-sensitivity}.

\subsection[Statistical summaries and comparisons]{Statistical summaries and comparisons}
\label{sec:statistical-summaries}
\label{sec:quantitative-comparisons}

For an endpoint comparison between groups $a$ and $b$, let $\overline z_a$ and $\overline z_b$ denote their mean task-oriented endpoints, with the appropriate aggregation for that comparison. Our primary endpoint effect size is the signed, unstandardized difference
$\Delta_{a-b}=\overline z_a-\overline z_b$.
We report $\Delta_{a-b}$ in $\mathrm{mNm}$ and convert the same difference to percentage points using $d_{\mathrm{nc}}$. The first column of \cref{tab:comparison-summary} gives the subtraction order.

For the main training, model-size, and flow-removal comparisons, we independently resample whole repetition endpoints within each condition and form a percentile-bootstrap interval from $\PFFProtocolBootstrapResamplesDisplay$ resamples. The main comparison has $\PFFProtocolMainRepetitions{}$ repetitions per condition. The two single-configuration ablations include conditions with only $\PFFProtocolDreamerAdditionalRepetitions{}$ repetitions, so their intervals are descriptive uncertainty summaries for the tested runs.

For the observation-family comparison, we average repetitions within each observation configuration before giving the four configurations equal weight within the supplied and withheld families. The configurations form four matched pairs that differ only in whether they include the dense flow field estimate. Its nested bootstrap resamples these pairs and then resamples whole training repetitions within each selected configuration. These intervals summarize the supplied-versus-withheld contrast across the four tested configuration pairs; they should not be extrapolated to other observation designs.

All resampling schemes retain the training repetition as a whole; they never treat episodes or training minutes as independent observations. Available incomplete records enter without imputation, as documented in \cref{appendix:experimental-accounting}.

For the observation-family trajectory summaries, we cap the mean time to first attainment at \PFFProtocolTrainingBudgetMinutes{} minutes: a repetition that does not attain contributes \PFFProtocolTrainingBudgetMinutes{} minutes. The capped time and between-configuration dispersion give the four configurations equal weight; lower capped times denote earlier first attainment. We time-average the root-mean-square deviation among the four configuration-mean curves over the common recorded window within each task to obtain the dispersion.
Trajectory summaries are reported in \cref{tab:observation-family-trajectories}.

\subsection[Initial drag-decrease analysis]{Initial drag-decrease analysis}
\label{sec:methods:inverse-response}

We characterize the physical response without orienting it by task or selecting trajectories by their eventual outcome. For filtered-drag sample $d_i$, let $g_i=d_i-d_0$, so that $g_i<0$ always denotes a drag decrease. We calculate a $\PFFProtocolInverseLocalAverageWindowS\,\mathrm{s}$ moving average and search all complete windows contained in the first $\PFFProtocolInverseInitialSearchWindowS\,\mathrm{s}$. An initial decrease is material when the minimum moving average is at most $-\PFFProtocolMaterialEffectThresholdMNm\,\mathrm{mNm}$. Separately, we call an increase sustained when the mean over the final $\PFFProtocolInverseSustainedResponseWindowS\,\mathrm{s}$ reaches $\PFFProtocolInverseSustainedDragIncreaseReferencePercent\%$ of the drag increase produced by the selected high-performing drag-maximizing sinusoid (a $\PFFPaperInverseSustainedDragIncreaseThresholdPercent\%$ drag increase relative to no control).

The primary analysis uses the $\PFFProtocolMaterialEffectThresholdMNm\,\mathrm{mNm}$ \hyperref[sec:uncertainties]{material-decrease threshold}. As a sensitivity analysis, we compare this threshold with the alternative fractions listed in \cref{tab:negative-response-material-threshold-sensitivity}. We define these fractions relative to the drag-reduction magnitude produced by the selected high-performing sinusoid for minimization and hold the sustained-increase criterion at \PFFProtocolInverseSustainedDragIncreaseReferencePercent\% of the selected high-performing sinusoid for maximization.
The corresponding results are reported in \cref{appendix:negative-response-analysis}.

\subsection[Observation-set ablation protocol]{Observation-set ablation protocol}
\label{sec:methods:observations}

We complement the \hyperref[sec:experiments:closed-loop]{main training experiment} by varying the observations that we supply to the agent.
We use the same physical setup, \texttt{DreamerV3} architecture, reward specification, and $\PFFProtocolTrainingBudgetMinutes$-minute training budget as in that experiment. The main configurations contribute the \PFFProtocolMainRepetitions{} repetitions from that experiment. We assign \PFFProtocolDreamerAdditionalRepetitions{} repetitions to each of the \PFFProtocolDreamerAdditionalConfigurationsPerObjective{} additional configurations per objective ($\PFFProtocolTrainingEpisodes$ episodes and $\PFFProtocolTrainingStepsDisplay$ environment steps per repetition). The drag measurement belongs to every observation set.

When we withhold flow observations, the agent receives the drag measurement, either alone or together with one or more of the following: a normalized episode-time index
\(t = \frac{\text{steps in episode}}{\text{horizon length}}\), the commanded rotation rate and rotation-rate feedback, or all of these.
When we supply flow observations, the agent additionally receives the dense flow field estimate.
The main comparison groups configurations by whether we supply or withhold the dense flow field estimate.

At each training episode, we first average the episode-average filtered-drag change across the available repetitions of each configuration. We then group configurations by whether we supply or withhold flow observations and compute the mean, minimum, maximum, and quartiles across the tested configurations. \Cref{fig:observations-all} reports these summaries.

\subsection[Model-size ablation protocol]{Model-size ablation protocol}
\label{sec:methods:model-size}

We study the effect of the model size of \texttt{DreamerV3} on the learning performance.
 We train the \texttt{DreamerV3} model with $\PFFProtocolSmallModelParametersMillions$ million parameters \PFFProtocolDreamerModelSizeRepetitions{} times on the drag-maximization task, using as observations the drag measurement, the commanded rotation rate, and its feedback from the motor.
 We compare it to the $\PFFProtocolStandardModelParametersMillions$-million-parameter model in the main experiment.
The corresponding learning curves are reported in \cref{fig:ablations:architecture}.

\subsection[Flow-withholding ablation protocol]{Flow-withholding ablation protocol}
\label{sec:methods:flow-withholding}

We compare two implementations for withholding flow observations from the $\PFFProtocolStandardModelParametersMillions$-million-parameter \texttt{DreamerV3} model.
We train the model \PFFProtocolDreamerZeroedFlowRepetitions{} times on the drag-maximization task using the drag measurement, commanded rotation rate, and motor rotation-rate feedback while fixing the flow-observation input to zero.
We compare this configuration with the one in the main experiment, where we omit the flow-observation branch.
The corresponding learning curves are reported in \cref{fig:ablations:architecture}.

\subsection[Learning-algorithm ablation protocol]{Learning-algorithm ablation protocol}
\label{sec:methods:algorithms}

We assess whether the drag-maximization outcome when we withhold flow observations is specific to \texttt{DreamerV3} by testing model-free on-policy and off-policy algorithms.
We consider the model-free on-policy \gls{ppo} algorithm \citep{schulman2017proximalpolicyoptimizationalgorithms} with both a feedforward network and a recurrent variant, \gls{ppor}, and the model-free off-policy \gls{sac} algorithm \citep{haarnoja2018softactorcriticoffpolicymaximum}.
We implement \gls*{ppo} and \gls*{sac} with \texttt{Stable-Baselines3} \citep{raffin2021stable} and \gls*{ppor} with \texttt{SB3-Contrib}, using version \PFFProtocolAlgorithmLibraryVersion{} of both libraries. We retain the libraries' default hyperparameters, except that we set the discount factor to $\gamma=\PFFProtocolAlgorithmDiscountFactor$ to account for the task horizon.
The learning times in \nameref{sec:results} motivate a budget of $\PFFProtocolAlgorithmBudgetMinutes$ minutes. In every case, we provide the temporal-progress observation in addition to the observations in \nameref{sec:results}, because not all implementations use recurrent networks. We run each algorithm in \PFFProtocolAlgorithmRepetitions{} independent project-level repetitions, each containing \PFFProtocolAlgorithmEpisodes{} physical episodes of $\PFFProtocolEpisodeDurationS\,\mathrm{s}$.
The corresponding learning curves are reported in \cref{fig:ablations:rl}.

\subsection[Online execution and replay analysis]{Online execution and replay analysis}
\label{sec:methods:replay}

At each checkpoint, we compare the task-oriented online episode score defined in \nameref{appendix:headline-training-metrics} with the mean score of its $\PFFProtocolReplayRepeats{}$ physical replays. We first average these replay scores over all $\PFFProtocolReplayCheckpoints{}$ checkpoints within each of the $\PFFProtocolReplaySeeds{}$ independent training repetitions. For the paired comparison, we average the checkpoint-wise online-minus-replay differences within each repetition. 

We use two-sided $\PFFProtocolConfidenceLevelPercent\%$ Student $t$ intervals with $\PFFProtocolReplayStudentTDegreesFreedom{}$ degrees of freedom for paired differences across the $\PFFProtocolReplaySeeds{}$ independent repetitions. With this number of repetitions, the normality assumption cannot be assessed reliably, so the intervals are descriptive uncertainty summaries.

We next compare the within-episode temporal response. During online execution, current observations can change the commanded rotation rate; during replay, we prescribe the recorded command trajectory. The logged replay commands closely match, but are not samplewise identical to, the online commands because of a deterministic numerical conversion that shifts some values by \PFFPaperTemporalCommandNumericalConversionShiftRPM{} revolution per minute and a brief startup alignment offset. For both tasks, their mean spectral divergence remains below \(\PFFPaperTemporalCommandSpectrumDivergenceUpperBoundBits\,\mathrm{bit}\). We therefore analyze only the measured motor rotation rate and filtered drag. Each normalized power spectrum is a Welch estimate based on \PFFProtocolTemporalWelchSegments{} segments of \PFFProtocolTemporalPrimarySegmentSeconds{} seconds with half-segment overlap, segment-wise mean removal, a periodic Hann window, and no zero padding \citep{welch1967power}. We omit the zero-frequency bin and normalize the remaining power to sum to one, so the comparison concerns spectral shape rather than signal mean, total power, or phase. At each checkpoint, we hold out each replay in turn and calculate the leave-one-out mean replay spectrum from the remaining replay spectra. We compute separate base-2 Jensen--Shannon divergences \citep{lin1991divergence} for the online and held-out replay spectra relative to that mean spectrum, then average each divergence over the held-out replays. The online spectrum and all \PFFProtocolReplayRepeats{} replay spectra must be defined at every checkpoint; the analysis does not average over a reduced replay set.

We first average values across checkpoints within each training repetition and then report the mean $\pm$ sample standard deviation across \PFFProtocolReplaySeeds{} independent repetitions. The paired difference subtracts among-replay dispersion from online--replay divergence within each repetition and is reported as its mean with a two-sided Student $t$ interval. As a sensitivity analysis, we replace isolated measured motor rotation-rate samples with magnitude above \PFFProtocolTemporalMotorFeedbackSpikeThresholdRPM{} revolutions per minute by within-trajectory linear interpolation between their adjacent unaffected samples. 
The paired performance comparisons and sensitivity results are reported in \cref{appendix:temporal-structure}, and the spectral comparisons in \cref{tab:temporal-structure}.

\clearpage
\section{Additional quantitative results}
\label{appendix:quantitative-analysis}
Definitions and statistical procedures are given in \nameref{appendix:headline-training-metrics}, \nameref{sec:statistical-summaries}, and \nameref{sec:methods:inverse-response} in the main manuscript.

\subsection{Training outcomes and threshold sensitivity}
\begin{papertable}
\centering
\footnotesize
\setlength{\tabcolsep}{2pt}
\caption{Main training outcomes. First-attainment entries are means $\pm$ sample standard deviations when every repetition reaches \PFFProtocolReferenceAttainmentPercent\% of the mean drag-change magnitude produced by the selected high-performing sinusoid for its task; otherwise, the table reports the attainment count. Final endpoints are means $\pm$ sample standard deviations across repetitions.}
\label{tab:headline-training-summary}
\begingroup\begin{tabular*}{\linewidth}{@{\extracolsep{\fill}}V{>{\raggedright\arraybackslash}p{.18\linewidth}}V{>{\raggedright\arraybackslash}p{.12\linewidth}}V{>{\raggedleft\arraybackslash}p{.12\linewidth}}V{>{\raggedleft\arraybackslash}p{.17\linewidth}}*{2}{V{>{\raggedleft\arraybackslash}p{.15\linewidth}}}@{}}
\toprule
Task & Flow observations & Repetitions & First-attainment time (min) & Final endpoint (mNm) & Final endpoint (\%) \\
\midrule
\ifdefined\PFFPaperNumbersLoaded\else
  \PackageError{paper_numbers}{Load paper_numbers.tex before headline_summary.tex}{The generated fragment requires the centralized paper-number registry.}
  \expandafter 
\fi
{Drag maximization} & {Supplied} & {10} & {$10.1\pm5.5$} & {$ 22.5\pm0.8 $} & {$ 25.5\pm0.9 $} \\
{Drag maximization} & {Withheld} & {10} & {Not attained ($0/10$)} & {$ 1.6\pm3.8 $} & {$ 1.8\pm4.3 $} \\
{Drag minimization} & {Supplied} & {10} & {$4.4\pm1.5$} & {$ 28.6\pm1.4 $} & {$ 32.4\pm1.6 $} \\
{Drag minimization} & {Withheld} & {10} & {$9.0\pm3.1$} & {$ 27.7\pm1.8 $} & {$ 31.4\pm2.1 $} \\

\bottomrule
\end{tabular*}\endgroup
\end{papertable}

The training contrast is unchanged across the tested thresholds. In \cref{tab:headline-training-threshold-sensitivity}, every supplied-flow repetition and every minimization repetition crosses each tested fraction of the selected high-performing sinusoid, whereas none of the withheld-flow maximization repetitions does.

\begin{papertable}
\centering
\scriptsize
\setlength{\tabcolsep}{2pt}
\caption{Sensitivity of first-attainment time to the required fraction of the selected high-performing sinusoid. The threshold columns give both the fraction \(\alpha\) and its task-specific drag-change equivalents relative to no control. Times are means $\pm$ sample standard deviations in active-interaction minutes when all $\PFFProtocolMainRepetitions{}$ repetitions attain the threshold. For the Max., withheld case, the dash indicates that the threshold was not attained in any repetition.}
\label{tab:headline-training-threshold-sensitivity}
\begingroup\begin{tabular*}{\linewidth}{@{\extracolsep{\fill}}*{7}{V{>{\centering\arraybackslash}p{.12\linewidth}}}@{}}
\toprule
\multicolumn{3}{c}{Threshold} & \multicolumn{4}{c}{First-attainment time (min)} \\
\cmidrule(lr){1-3}\cmidrule(lr){4-7}
\(\alpha\) (\%) & Max. drag increase (\%) & Min. drag reduction (\%) & Max., supplied & Max., withheld & Min., supplied & Min., withheld \\
\midrule
\ifdefined\PFFPaperNumbersLoaded\else
  \PackageError{paper_numbers}{Load paper_numbers.tex before learning_threshold_sensitivity.tex}{The generated fragment requires the centralized paper-number registry.}
  \expandafter 
\fi
{80} & {21.3} & {23.7} & {$10.1\pm5.5$} & {---} & {$4.4\pm1.5$} & {$9.0\pm3.1$} \\
{85} & {22.6} & {25.2} & {$11.2\pm5.5$} & {---} & {$4.7\pm1.6$} & {$9.7\pm3.3$} \\
{90} & {24.0} & {26.7} & {$15.4\pm7.0$} & {---} & {$5.7\pm1.4$} & {$10.2\pm2.9$} \\
{95} & {25.3} & {28.2} & {$20.9\pm11.7$} & {---} & {$6.5\pm2.1$} & {$10.6\pm2.8$} \\
{100} & {26.6} & {29.7} & {$24.1\pm11.0$} & {---} & {$7.4\pm2.5$} & {$13.2\pm4.2$} \\

\bottomrule
\end{tabular*}\endgroup
\end{papertable}

\subsection{Endpoint comparisons}
\begin{papertable}
\centering
\footnotesize
\setlength{\tabcolsep}{3pt}
\caption{Endpoint comparisons. The first column gives the subtraction order, and the signed difference in $\mathrm{mNm}$ is the primary effect size. Intervals use the percentile bootstrap.}
\label{tab:comparison-summary}
\begingroup\begin{tabular*}{\linewidth}{@{\extracolsep{\fill}}V{>{\raggedright\arraybackslash}m{0.254\linewidth}}V{>{\centering\arraybackslash}m{0.19\linewidth}}V{>{\centering\arraybackslash}m{0.16\linewidth}}V{>{\centering\arraybackslash}m{0.182\linewidth}}V{>{\centering\arraybackslash}m{0.15\linewidth}}@{}}
\toprule
Comparison & Independent units & Primary effect size (mNm) & Difference (percentage points) & \PFFProtocolConfidenceLevelPercent\% interval (mNm) \\
\midrule
\ifdefined\PFFPaperNumbersLoaded\else
  \PackageError{paper_numbers}{Load paper_numbers.tex before comparisons.tex}{The generated fragment requires the centralized paper-number registry.}
  \expandafter 
\fi
{Main max.:\linebreak supplied $-$ withheld} & {10+10 training runs} & {$21.0$} & {$23.7$} & {$[19.0,23.5]$} \\
{Main min.:\linebreak supplied $-$ withheld} & {10+10 training runs} & {$0.9$} & {$1.0$} & {$[-0.5,2.2]$} \\
{Observation families, max.:\linebreak supplied $-$ withheld} & {4 matched configuration pairs} & {$21.4$} & {$24.2$} & {$[16.4,26.9]$} \\
{Observation families, min.:\linebreak supplied $-$ withheld} & {4 matched configuration pairs} & {$3.1$} & {$3.5$} & {$[0.5,6.9]$} \\
{Model size, max.:\linebreak 1 million $-$ 25 million} & {3+10 training runs} & {$-9.4$} & {$-10.7$} & {$[-11.5,-6.9]$} \\
{Flow removal, max.:\linebreak zeroed $-$ omitted} & {3+10 training runs} & {$1.9$} & {$2.1$} & {$[-6.7,10.2]$} \\

\bottomrule
\end{tabular*}\endgroup
\end{papertable}
\subsection[Prevalence of the initial drag decrease]{Prevalence of the initial drag decrease}
\label{appendix:negative-response-analysis}

The initial material drag decrease is prevalent during online execution and open-loop replay. It occurs in \PFFPaperNegativeOnlineMaterialPercent\% of the \PFFPaperNegativeOnlineEligible{} online \texttt{DreamerV3} trajectories that we analyze and in all \PFFPaperNegativePairedReplayEligible{} pointwise replay means. Among records that sustain a drag increase, the decrease occurs first in \PFFPaperNegativeOnlineSustainedPositivePrecededByMaterialNegativePercent\% of the \PFFPaperNegativeOnlineSustainedPositive{} online trajectories and in all \PFFPaperNegativePairedReplaySustainedPositive{} replay means. Each replay mean is calculated pointwise from \PFFProtocolReplayRepeats{} physical trials of one recorded action trajectory. These frequencies describe the analyzed records; they are not estimates of population probabilities, and the records are not independent training repetitions. Thus, under matched operating conditions, prescribing the recorded command sequences reproduces the transient without online policy correction.

\begin{papertable}
\centering
\footnotesize
\setlength{\tabcolsep}{2.5pt}
\caption{Sensitivity of initial-decrease prevalence and decrease-before-increase ordering to the material-decrease threshold. Percentage thresholds are fractions of the drag-reduction magnitude produced by the selected high-performing sinusoid for minimization. Initial-decrease entries use all eligible records; ordering entries condition on records with a sustained increase. The sustained-increase criterion remains \PFFProtocolInverseSustainedDragIncreaseReferencePercent\% of the selected high-performing sinusoid for maximization.}
\label{tab:negative-response-material-threshold-sensitivity}
\begingroup\begin{tabular*}{\linewidth}{@{\extracolsep{\fill}}V{>{\raggedright\arraybackslash}p{.18\linewidth}}V{>{\raggedright\arraybackslash}p{.17\linewidth}}V{>{\raggedright\arraybackslash}p{.21\linewidth}}V{>{\raggedright\arraybackslash}p{.17\linewidth}}V{>{\raggedright\arraybackslash}p{.21\linewidth}}@{}}
\toprule
Material-decrease threshold & Online initial decrease & Online decrease before sustained increase & Paired means initial decrease & Paired means decrease before sustained increase \\
\midrule
\ifdefined\PFFPaperNumbersLoaded\else
  \PackageError{paper_numbers}{Load paper_numbers.tex before negative_response_threshold_sensitivity.tex}{The generated fragment requires the centralized paper-number registry.}
  \expandafter 
\fi
{Primary: $1.000\,\mathrm{mNm}$} & {$4641/4918$ $(94.4\%)$} & {$1049/1092$ $(96.1\%)$} & {$360/360$ $(100.0\%)$} & {$161/161$ $(100.0\%)$} \\
{$5\%$ ($1.309\,\mathrm{mNm}$)} & {$4549/4918$ $(92.5\%)$} & {$1011/1092$ $(92.6\%)$} & {$360/360$ $(100.0\%)$} & {$161/161$ $(100.0\%)$} \\
{$10\%$ ($2.617\,\mathrm{mNm}$)} & {$4122/4918$ $(83.8\%)$} & {$792/1092$ $(72.5\%)$} & {$351/360$ $(97.5\%)$} & {$153/161$ $(95.0\%)$} \\
{$15\%$ ($3.926\,\mathrm{mNm}$)} & {$3806/4918$ $(77.4\%)$} & {$595/1092$ $(54.5\%)$} & {$321/360$ $(89.2\%)$} & {$123/161$ $(76.4\%)$} \\
{$20\%$ ($5.235\,\mathrm{mNm}$)} & {$3587/4918$ $(72.9\%)$} & {$460/1092$ $(42.1\%)$} & {$300/360$ $(83.3\%)$} & {$103/161$ $(64.0\%)$} \\

\bottomrule
\end{tabular*}\endgroup
\end{papertable}

Initial decreases remain present in most online trajectories at every tested threshold, whereas their prevalence among trajectories with a sustained increase declines as the required decrease magnitude increases. The paired-replay means show the same attenuation at the stricter thresholds.

The prescribed-input controls delimit this observation. An initial material decrease occurs in \PFFPaperNegativeSinusoidalWithFlowMaterialPercent\% of the \PFFPaperNegativeSinusoidalWithFlowEligible{} open-loop grid condition means with incoming flow and in all \PFFPaperNegativeSinusoidalHigherFrequencyEligible{} higher-frequency sinusoidal condition means, but in none of the \PFFPaperNegativeSinusoidalWithoutFlowEligible{} corresponding condition means \hyperref[sec:water-channel:torque]{without incoming flow}. Each prescribed-input record is a condition mean over its available trials. The absence of the decrease without incoming flow is inconsistent with a comparable motor-only artifact in these controls, while its incomplete prevalence across the main grid shows that the response depends on the applied trajectory.

The decrease also occurs in \PFFPaperNegativeZeroedFlowMaterialPercent\% of the \PFFPaperNegativeZeroedFlowEligible{} online trajectories with the flow-observation input zeroed and in all \PFFPaperNegativeSmallModelEligible{} trajectories from the one-million-parameter \texttt{DreamerV3} configuration. These two groups are subsets of the online \texttt{DreamerV3} cohort, not additional records. It likewise occurs in \PFFPaperNegativeAlgorithmBaselinesMaterialPercent\% of the \PFFPaperNegativeAlgorithmBaselinesEligible{} \PFFProtocolAlgorithmBudgetMinutes-minute trajectories from the \gls*{ppo}, \gls*{ppor}, and \gls*{sac} experiments; this count excludes the extended \PFFProtocolExtendedSACHours-hour \gls*{sac} run.

When an initial material drag decrease precedes a sustained increase, the ordered response constitutes a trajectory-level inverse response. The initial decrease favors minimization but is adverse to maximization. Rotation-induced drag reduction and accompanying changes in separation and wake topology provide qualitative precedent for the physical decrease \citep{aljure2015influence}; unsteady, rotation-dependent force and wake evolution provide further context \citep{badr1990unsteady}. Neither study identifies the mechanism in this confined apparatus. The transient may therefore contribute to the maximization-specific discovery difficulty, but these descriptive observations neither prove a unique cause of the learning asymmetry nor classify the full nonlinear plant as globally non-minimum-phase.

\clearpage
\section{Ablations}
\label{sec:ablations}
We report ablations of (i) the observation sets (\cref{sec:ablations:observations-sets}), (ii) the model size (\cref{sec:ablations:model-size}), (iii) how we withhold flow observations (\cref{sec:ablations:observations}), and (iv) the learning algorithm (\cref{appendix:rl}).

\subsection{Ablations on the observation sets}
\label{sec:ablations:observations-sets}
The protocol is described in \nameref{sec:methods:observations} in the main manuscript.

The trends differ sharply between the tasks. In drag maximization, configurations without flow observations show little systematic improvement, whereas those with flow observations consistently reach higher performance. In drag minimization, both families learn high-performing policies, although configurations without flow observations exhibit greater variability.
These curves differ from those in \nameref{sec:experiments:closed-loop}, which reports variability across repetitions of a single configuration. Here, the summaries describe the mean and variability across the tested configurations after averaging repetitions within each configuration.
When the agent observes only drag, the weaker drag-minimization outcome is consistent with state aliasing \citep{whitehead1991learning}. These configurations contribute to the aggregate statistics and lower the family-level curve.
Across these configurations, supplying flow observations is particularly helpful for discovering high-performing drag-maximizing policies; its effect on drag minimization is smaller and appears mainly as reduced variability.

\PFFFigure{S1}

\subsection{Ablations on the model size}
\label{sec:ablations:model-size}
The protocol is described in \nameref{sec:methods:model-size} in the main manuscript.

\Cref{fig:ablations:architecture} shows the same drag-maximization outcome for $\PFFProtocolStandardModelParametersMillions$-million- and $\PFFProtocolSmallModelParametersMillions$-million-parameter models: when we withhold flow observations, neither model discovers a high-performing policy.

\PFFFigure{S2}

\subsection{Implementations for withholding flow observations}
\label{sec:ablations:observations}
The protocol is described in \nameref{sec:methods:flow-withholding} in the main manuscript.

\Cref{fig:ablations:architecture} shows similar performance for the two implementations. One zero-valued-input repetition performs slightly better on a few rollouts, but neither implementation reaches a high-performing policy.
Because the two implementations produce similar outcomes in these trials, elsewhere we omit the flow-observation branch whenever we withhold flow observations.

\subsection{Ablations on the learning algorithm}
\label{appendix:rl}
The protocol is described in \nameref{sec:methods:algorithms} in the main manuscript.

\Cref{fig:ablations:rl} reports the drag-maximization learning curves. None of the \gls*{ppo}, \gls*{ppor}, or \gls*{sac} configurations without flow observations reaches a high-performing policy; the extended \gls*{sac} run remains unsuccessful after \PFFProtocolExtendedSACHours{} hours. When we supply flow observations, \texttt{DreamerV3} discovers a high-performing policy within minutes; see \nameref{sec:results}.

\PFFFigure{S3}

\subsection[Quantitative endpoint and trajectory comparisons]{Quantitative endpoint and trajectory comparisons}
\label{appendix:ablation-statistics}

At this endpoint, the supplied-minus-withheld family difference is $\PFFPaperObservationMaxDifferenceMNm\,\mathrm{mNm}$ ($\PFFPaperObservationMaxDifferencePercentagePoints$ percentage points; $\PFFProtocolConfidenceLevelPercent\%$ confidence interval $[\PFFPaperObservationMaxConfidenceLowerMNm,\PFFPaperObservationMaxConfidenceUpperMNm]\,\mathrm{mNm}$) for drag maximization and $\PFFPaperObservationMinDifferenceMNm\,\mathrm{mNm}$ ($\PFFPaperObservationMinDifferencePercentagePoints$ percentage points; $\PFFProtocolConfidenceLevelPercent\%$ confidence interval $[\PFFPaperObservationMinConfidenceLowerMNm,\PFFPaperObservationMinConfidenceUpperMNm]\,\mathrm{mNm}$) for drag minimization. Reducing the model from $\PFFProtocolStandardModelParametersMillions$ million to $\PFFProtocolSmallModelParametersMillions$ million parameters changes the drag-maximization endpoint by $\PFFPaperModelSizeMaxDifferenceMNm\,\mathrm{mNm}$ ($\PFFPaperModelSizeMaxDifferencePercentagePoints$ percentage points; $\PFFProtocolConfidenceLevelPercent\%$ confidence interval $[\PFFPaperModelSizeMaxConfidenceLowerMNm,\PFFPaperModelSizeMaxConfidenceUpperMNm]\,\mathrm{mNm}$). Fixing the flow-observation input to zero rather than omitting the flow-observation branch changes it by $\PFFPaperFlowRemovalMaxDifferenceMNm\,\mathrm{mNm}$ ($\PFFPaperFlowRemovalMaxDifferencePercentagePoints$ percentage points; $\PFFProtocolConfidenceLevelPercent\%$ confidence interval $[\PFFPaperFlowRemovalMaxConfidenceLowerMNm,\PFFPaperFlowRemovalMaxConfidenceUpperMNm]\,\mathrm{mNm}$). \nameref{sec:quantitative-comparisons} specifies the effect-size definitions, subtraction order, and interval procedures. The intervals are wide because each additional ablation configuration has \PFFProtocolDreamerAdditionalRepetitions{} repetitions.

\begin{papertable}
\centering
\footnotesize
\setlength{\tabcolsep}{2pt}
\caption{Observation-family trajectory summaries over the \PFFProtocolTrainingBudgetMinutes-minute budget. First attainment and final-window attainment require \PFFProtocolReferenceAttainmentPercent\% of the drag-change magnitude produced by the selected high-performing sinusoid for the corresponding task: a \PFFPaperHeadlineMaxAttainmentDragChangePercent\% drag increase for maximization and a \PFFPaperHeadlineMinAttainmentDragChangePercent\% drag reduction for minimization, relative to no control. Final-window attainment means that the endpoint defined in \nameref{appendix:headline-training-metrics} meets the corresponding criterion. Counts are raw repetition counts. We cap the mean time to first attainment at \PFFProtocolTrainingBudgetMinutes{} minutes: a repetition that does not attain contributes \PFFProtocolTrainingBudgetMinutes{} minutes. The capped time and between-configuration dispersion give the four configurations equal weight; lower capped times denote earlier first attainment. We time-average the root-mean-square deviation among the four configuration-mean curves over the common recorded window within each task to obtain the dispersion. These descriptive summaries complement the primary endpoint effects in \cref{tab:comparison-summary}.}
\label{tab:observation-family-trajectories}
\begingroup\begin{tabular*}{\linewidth}{@{\extracolsep{\fill}}V{>{\raggedright\arraybackslash}p{.15\linewidth}}V{>{\raggedright\arraybackslash}p{.12\linewidth}}*{2}{V{>{\centering\arraybackslash}p{.13\linewidth}}}V{>{\centering\arraybackslash}p{.20\linewidth}}V{>{\centering\arraybackslash}p{.16\linewidth}}@{}}
\toprule
Task & Flow observations & First attained (runs) & Final window (runs) & Mean time to first attainment, capped at \PFFProtocolTrainingBudgetMinutes{} min & Mean dispersion (mNm) \\
\midrule
\ifdefined\PFFPaperNumbersLoaded\else
  \PackageError{paper_numbers}{Load paper_numbers.tex before observation_trajectory_summary.tex}{The generated fragment requires the centralized paper-number registry.}
  \expandafter 
\fi
{Drag maximization} & {Supplied} & {19/19} & {19/19} & {7.9} & {1.30} \\
{Drag maximization} & {Withheld} & {1/19} & {0/19} & {57.6} & {3.53} \\
{Drag minimization} & {Supplied} & {19/19} & {19/19} & {4.7} & {1.57} \\
{Drag minimization} & {Withheld} & {18/19} & {16/19} & {21.4} & {4.46} \\

\bottomrule
\end{tabular*}\endgroup
\end{papertable}

Together, \cref{tab:comparison-summary,tab:observation-family-trajectories} distinguish the endpoint from first-attainment timing and between-configuration trajectory dispersion. For drag maximization, supplying flow observations raises the endpoint, yields first attainment in all \PFFPaperObservationMaxFlowRunCount{} repetitions rather than \PFFPaperObservationMaxNoFlowAttainedCount{} of \PFFPaperObservationMaxNoFlowRunCount{}, and yields final-window attainment in all repetitions rather than none. The configuration-balanced withheld-minus-supplied difference in mean time to first attainment, capped at \PFFProtocolTrainingBudgetMinutes{} minutes, is \PFFPaperObservationMaxRestrictedTimeDifferenceMinutes{} minutes ($\PFFProtocolConfidenceLevelPercent\%$ confidence interval $[\PFFPaperObservationMaxRestrictedTimeConfidenceLowerMinutes,\PFFPaperObservationMaxRestrictedTimeConfidenceUpperMinutes]$ minutes). For drag minimization, the endpoint difference is smaller, but supplying flow observations yields earlier first attainment in the matched observation designs and final-window attainment in all \PFFPaperObservationMinFlowRunCount{} repetitions rather than \PFFPaperObservationMinNoFlowFinalWindowAttainedCount{} of \PFFPaperObservationMinNoFlowRunCount{}; the corresponding capped-time difference is \PFFPaperObservationMinRestrictedTimeDifferenceMinutes{} minutes ($\PFFProtocolConfidenceLevelPercent\%$ confidence interval $[\PFFPaperObservationMinRestrictedTimeConfidenceLowerMinutes,\PFFPaperObservationMinRestrictedTimeConfidenceUpperMinutes]$ minutes). Configurations in which we supply flow observations have dispersion point estimates that are lower by \PFFPaperObservationMaxDispersionDifferenceMNm{} and \PFFPaperObservationMinDispersionDifferenceMNm{} $\mathrm{mNm}$ for maximization and minimization, respectively; their descriptive intervals are $[\PFFPaperObservationMaxDispersionConfidenceLowerMNm,\PFFPaperObservationMaxDispersionConfidenceUpperMNm]$ and $[\PFFPaperObservationMinDispersionConfidenceLowerMNm,\PFFPaperObservationMinDispersionConfidenceUpperMNm]$ $\mathrm{mNm}$. Because only four matched configuration pairs define these family contrasts, the intervals do not support generalization beyond the tested designs. The smaller model has a lower drag-maximization endpoint than the standard model, although neither model without flow observations reaches \PFFProtocolReferenceAttainmentPercent\% of the drag increase produced by the selected high-performing sinusoid for maximization. In the zero-valued-flow-input ablation, first attainment of that level occurs in \PFFPaperZeroedFlowAttainedCount{} of \PFFPaperZeroedFlowRunCount{} repetitions after \PFFPaperZeroedFlowFirstAttainmentMinute{} minutes, but final-window attainment occurs in \PFFPaperZeroedFlowFinalWindowAttainedCount{} of \PFFPaperZeroedFlowRunCount{} repetitions; this isolated crossing therefore does not establish a reliably high-performing outcome. Its endpoint interval includes zero and material differences in either direction, so these trials also do not establish endpoint equivalence with omission of the flow-observation branch.

\clearpage
\section[Online execution and open-loop replay]{Online execution and open-loop replay}
\label{appendix:temporal-structure}

We compare mean episode performance and within-episode responses under online execution and open-loop replay. The analysis protocol is given in \nameref{sec:methods:replay} in the main manuscript.

Across repetitions, replay produces a $\PFFPaperReplayMaxAllCheckpointPercentPM\%$ drag increase for maximization and a $\PFFPaperReplayMinAllCheckpointPercentPM\%$ drag reduction for minimization, relative to the no-control mean; the plus--minus terms are sample standard deviations across repetitions. The resulting difference is $\PFFPaperReplayMaxDifferenceMNmPM\,\mathrm{mNm}$ for drag maximization and $\PFFPaperReplayMinDifferenceMNmPM\,\mathrm{mNm}$ for drag minimization; these plus--minus terms are likewise sample standard deviations across repetitions. The corresponding two-sided $\PFFProtocolConfidenceLevelPercent\%$ Student $t$ intervals with $\PFFProtocolReplayStudentTDegreesFreedom{}$ degrees of freedom are $[\PFFPaperReplayMaxConfidenceLowerMNm,\PFFPaperReplayMaxConfidenceUpperMNm]\,\mathrm{mNm}$ and $[\PFFPaperReplayMinConfidenceLowerMNm,\PFFPaperReplayMinConfidenceUpperMNm]\,\mathrm{mNm}$, respectively. With only $\PFFProtocolReplaySeeds{}$ repetitions, the normality assumption cannot be assessed reliably, so these intervals are descriptive uncertainty summaries; an interval containing zero does not establish equivalence.

\begin{papertable}
\centering
\footnotesize
\setlength{\tabcolsep}{2pt}
\caption{Base-2 Jensen--Shannon divergence between normalized spectra of the measured motor rotation rate and filtered drag. For each held-out replay, online--replay divergence compares the online spectrum with the mean of the remaining replay spectra, whereas among-replay dispersion compares the held-out replay spectrum with that same mean; both quantities average over held-out replays. The scale is zero for identical spectra and reaches one bit only for disjoint spectral support. We first average values across checkpoints within each training repetition and then report the mean $\pm$ sample standard deviation across \PFFProtocolReplaySeeds{} independent repetitions. The paired difference subtracts among-replay dispersion from online--replay divergence within each repetition and is reported as its mean with a two-sided Student $t$ interval.}
\label{tab:temporal-structure}
\begingroup\begin{tabular*}{\linewidth}{@{\extracolsep{\fill}}V{>{\raggedright\arraybackslash}p{.11\linewidth}}V{>{\raggedright\arraybackslash}p{.18\linewidth}}*{2}{V{>{\centering\arraybackslash}p{.19\linewidth}}}V{>{\centering\arraybackslash}p{.27\linewidth}}@{}}
\toprule
Task & Response & Online--replay divergence (bit) & Among-replay dispersion (bit) & Paired difference\newline (bit; \PFFProtocolConfidenceLevelPercent\% interval) \\
\midrule
\ifdefined\PFFPaperNumbersLoaded\else
  \PackageError{paper_numbers}{Load paper_numbers.tex before temporal_summary.tex}{The generated fragment requires the centralized paper-number registry.}
  \expandafter 
\fi
{Maximize} & {Motor rotation rate} & {$0.017\pm0.010$} & {$0.0158\pm0.0028$} & {$0.0010\,[-0.0248,0.0268]$} \\
{} & {Filtered drag} & {$0.064\pm0.011$} & {$0.0217\pm0.0022$} & {$0.0425\,[0.0212,0.0638]$} \\
\addlinespace
{Minimize} & {Motor rotation rate} & {$0.0196\pm0.0046$} & {$0.0219\pm0.0067$} & {$-0.0024\,[-0.0298,0.0251]$} \\
{} & {Filtered drag} & {$0.0610\pm0.0027$} & {$0.0135\pm0.0020$} & {$0.0475\,[0.0416,0.0535]$} \\

\bottomrule
\end{tabular*}\endgroup
\end{papertable}

We retain the unmodified motor-spectrum estimates in \cref{tab:temporal-structure}. The interpolation sensitivity analysis described in \nameref{sec:methods:replay} applies to \PFFPaperTemporalMotorFeedbackOnlineInterpolatedSamples{} samples in \PFFPaperTemporalMotorFeedbackOnlineAffectedTrajectories{} of \PFFPaperTemporalOnlineTrajectories{} online trajectories and \PFFPaperTemporalMotorFeedbackReplayInterpolatedSamples{} samples in \PFFPaperTemporalMotorFeedbackReplayAffectedTrajectories{} of \PFFProtocolReplayTrajectoriesDisplay{} replay trajectories. For maximization, the online--replay divergence and among-replay dispersion decrease from \PFFPaperTemporalMaxMotorUnmodifiedSpectrumDivergenceBitsMean{} and \PFFPaperTemporalMaxMotorUnmodifiedReplayDispersionBitsMean{} to \PFFPaperTemporalMaxMotorInterpolatedSpectrumDivergenceBitsMean{} and \PFFPaperTemporalMaxMotorInterpolatedReplayDispersionBitsMean{} bits, respectively; for minimization, they decrease from \PFFPaperTemporalMinMotorUnmodifiedSpectrumDivergenceBitsMean{} and \PFFPaperTemporalMinMotorUnmodifiedReplayDispersionBitsMean{} to \PFFPaperTemporalMinMotorInterpolatedSpectrumDivergenceBitsMean{} and \PFFPaperTemporalMinMotorInterpolatedReplayDispersionBitsMean{} bits. Both paired point differences reverse sign, but both corresponding descriptive intervals continue to include zero, so the conclusion for motor rotation rate is unchanged; the filtered-drag spectra are not altered by this sensitivity analysis.

The descriptive paired intervals in \cref{tab:temporal-structure} exclude zero for filtered drag in both tasks and include zero for motor rotation rate.
Under matched and stabilized conditions, the recorded sequence reproduces comparable mean drag but not the full temporal structure of the drag response.

\clearpage
\section[\texorpdfstring{Experimental data accounting}{Experimental data accounting}]{Experimental data accounting}
\label{appendix:experimental-accounting}

\Cref{tab:experimental-accounting} reports nominal protocol time by collection. Each episode or trial contributes \PFFProtocolEpisodeDurationS{} seconds of active time and one prescribed pre-trial stabilization interval. Total time is the sum of these two components. We exclude variable settling, setup, calibration, maintenance, and interruptions because we did not log them uniformly; the values are therefore not elapsed wall-clock times.

\begin{papertable}
\centering
\scriptsize
\setlength{\tabcolsep}{1.5pt}
\caption{Nominal experimental accounting in hours.}
\label{tab:experimental-accounting}
\begingroup\begin{tabular*}{\linewidth}{@{\extracolsep{\fill}}V{>{\raggedright\arraybackslash}p{.22\linewidth}}V{>{\raggedright\arraybackslash}p{.20\linewidth}}V{>{\raggedleft\arraybackslash}p{.10\linewidth}}V{>{\raggedleft\arraybackslash}p{.12\linewidth}}V{>{\raggedleft\arraybackslash}p{.15\linewidth}}V{>{\raggedleft\arraybackslash}p{.12\linewidth}}@{}}
\toprule
Collection & Nominal design & Count & Active & Stabilization & Total \\
\midrule
\ifdefined\PFFPaperNumbersLoaded\else
  \PackageError{paper_numbers}{Load paper_numbers.tex before accounting_nominal.tex}{The generated fragment requires the centralized paper-number registry.}
  \expandafter 
\fi
{\texttt{DreamerV3}: main} & {40 runs $\times 60$} & {2,400} & {40.00} & {40.00} & {80.00} \\
{\texttt{DreamerV3}: observation sets} & {36 runs $\times 60$} & {2,160} & {36.00} & {36.00} & {72.00} \\
{\texttt{DreamerV3}: model size} & {3 runs $\times 60$} & {180} & {3.00} & {3.00} & {6.00} \\
{\texttt{DreamerV3}: zeroed-flow input} & {3 runs $\times 60$} & {180} & {3.00} & {3.00} & {6.00} \\
\midrule
{\texttt{DreamerV3} subtotal} & {82 runs $\times 60$} & {4,920} & {82.00} & {82.00} & {164.00} \\
\midrule
{Recorded-trajectory replay} & {2 objectives $\times 3 \times 60 \times 5$} & {1,800} & {30.00} & {30.00} & {60.00} \\
{Incoming-flow grid baseline} & {129 settings $\times 3$} & {387} & {6.45} & {6.45} & {12.90} \\
{No-incoming-flow torque baseline} & {129 settings $\times 3$} & {387} & {6.45} & {1.08} & {7.53} \\
{Higher-frequency baseline} & {29 trials} & {29} & {0.48} & {0.16} & {0.64} \\
{\gls*{ppo}} & {3 runs $\times 30$} & {90} & {1.50} & {1.50} & {3.00} \\
{\gls*{ppor}} & {3 runs $\times 30$} & {90} & {1.50} & {1.50} & {3.00} \\
{\gls*{sac}} & {3 runs $\times 30$} & {90} & {1.50} & {1.50} & {3.00} \\
{Extended \gls*{sac}} & {1 run $\times 600$} & {600} & {10.00} & {10.00} & {20.00} \\
\midrule
{\textbf{All collections}} & {} & {\textbf{8,393}} & {\textbf{139.88}} & {\textbf{134.19}} & {\textbf{274.07}} \\

\bottomrule
\end{tabular*}\endgroup
\par\smallskip
\begin{minipage}{.98\linewidth}
\footnotesize
 In the observation-sets collection, one final episode is absent and another contains \PFFPaperAccountingDreamerPartialObservationAblationSamples{} of \PFFProtocolSamplesPerEpisodeDisplay{} samples; analyses use the available records without imputation.
\end{minipage}
\end{papertable}

\end{appendices}
\end{document}